\documentclass{article}

\usepackage[final]{neurips_2023}

\usepackage[utf8]{inputenc} 
\usepackage[T1]{fontenc}    
\usepackage{hyperref}       
\usepackage{url}            
\usepackage{booktabs}       
\usepackage{amsfonts}       
\usepackage{nicefrac}       
\usepackage{microtype}      
\usepackage{xcolor}         

\RequirePackage{fancyhdr}
\RequirePackage{algorithm}
\RequirePackage{algorithmic}
\RequirePackage{eso-pic} 
\RequirePackage{forloop}
\usepackage{natbib}
\setcitestyle{aysep{}} 
\usepackage{graphicx}
\usepackage{amsmath}
\usepackage{amssymb}
\usepackage{mathtools}
\usepackage{amsthm}

\usepackage{cleveref}
\crefname{equation}{Eq.}{Eqs.}
\crefname{table}{Table}{Tables}
\crefname{figure}{Figure}{Figures}
\crefname{section}{Section}{Sections}
\crefname{algorithm}{Algorithm}{Algorithms}

\theoremstyle{plain}
\newtheorem{theorem}{Theorem}[section]

\theoremstyle{definition}
\newtheorem{definition}[theorem]{Definition}

\theoremstyle{remark}

\usepackage{xcolor,colortbl}
\usepackage{adjustbox}
\usepackage{url}
\usepackage{multirow,multicol,xspace}
\usepackage{float}
\usepackage{graphics}
\usepackage{paralist}
\usepackage{wrapfig}
\usepackage[normalem]{ulem}

\usepackage{pgfplots}
\pgfplotsset{width=8cm,compat=1.17} 
\usepackage{pifont}
\newcommand{\cmark}{\ding{51}}%
\newcommand{\xmark}{\ding{55}}%
\usepackage{subcaption,ragged2e}
\usepackage{caption}

\newcommand{\opfunc}[1]{\textsc{#1}}

\title{Data-Centric Learning from Unlabeled Graphs \\ with Diffusion Model}

\author{%
  Gang Liu \\
  University of Notre Dame\\
  \texttt{gliu7@nd.edu} \\
  \And
  Eric Inae \\
  University of Notre Dame\\
  \texttt{einae@nd.edu} \\
  \And
  Tong Zhao \\
  Snap Inc. \\
  \texttt{tzhao@snap.com} \\
  \And
  Jiaxin Xu \\
  University of Notre Dame\\
  \texttt{jxu24@nd.edu} \\
  \And
  Tengfei Luo \\
  University of Notre Dame\\
  \texttt{tluo@nd.edu} \\
  \And
  Meng Jiang \\
  University of Notre Dame\\
  \texttt{mjiang2@nd.edu} \\
}

\begin{document}
\maketitle

\newcommand{\method}{\textsc{DCT}\xspace}
\newcommand{\gin}{\textsc{GIN}\xspace}

\def\T{{\mathcal{T}}}
\def\ie{\textit{i.e., }}
\def\eg{\textit{e.g., }}
\def\dcc{black}

\newcommand{\infomax}{\textsc{InfoMax}\xspace}
\newcommand{\edgepred}{\textsc{EdgePred}\xspace}
\newcommand{\contextpred}{\textsc{ContextPred}\xspace}
\newcommand{\attrmask}{\textsc{AttrMask}\xspace}
\newcommand{\joao}{\textsc{JOAO}\xspace}
\newcommand{\graphlog}{\textsc{GraphLoG}\xspace}
\newcommand{\mgssl}{\textsc{MGSSL}\xspace}
\newcommand{\dsla}{\textsc{D-SLA}\xspace}

\newcommand{\infograph}{\textsc{InfoGraph}\xspace}
\newcommand{\streal}{\textcolor{\dcc}{\textsc{ST-Real}}\xspace}
\newcommand{\stgen}{\textcolor{\dcc}{\textsc{ST-Gen}}\xspace}

\newcommand{\grea}{\textcolor{\dcc}{\textsc{GREA}}\xspace}
\newcommand{\gmix}{\textcolor{\dcc}{\textsc{G-MixUp}}\xspace}
\newcommand{\flag}{\textcolor{\dcc}{\textsc{FLAG}}\xspace}

\newcommand{\glass}{{GlassTemp}\xspace}
\newcommand{\melting}{{MeltingTemp}\xspace}
\newcommand{\density}{{PolyDensity}\xspace}
\newcommand{\thermal}{{ThermCond}\xspace}
\newcommand{\oxygen}{{O$_2$Perm}\xspace}

\newcommand{\hiv}{{ogbg-HIV}\xspace}
\newcommand{\toxcast}{{ogbg-ToxCast}\xspace}
\newcommand{\toxt}{{ogbg-Tox21}\xspace}
\newcommand{\bace}{{ogbg-BACE}\xspace}
\newcommand{\bbbp}{{ogbg-BBBP}\xspace}
\newcommand{\clintox}{{ogbg-ClinTox}\xspace}
\newcommand{\sider}{{ogbg-SIDER}\xspace}
\newcommand{\freesolv}{{ogbg-FreeSolv}\xspace}
\newcommand{\lipo}{{ogbg-Lipo}\xspace}
\newcommand{\esol}{{ogbg-ESOL}\xspace}
\newcommand{\ppi}{{PPI}\xspace}

\def\sqPDF#1#2{0 0 m #1 0 l #1 #1 l 0 #1 l h}
\def\trianPDF#1#2{0 0 m #1 0 l #2 4.5 l h}
\def\uptrianPDF#1#2{#2 0 m #1 4.5 l 0 4.5 l h}
\def\circPDF#1#2{#1 0 0 #1 #2 #2 cm .1 w .5 0 m
   .5 .276 .276 .5 0 .5 c -.276 .5 -.5 .276 -.5 0 c
   -.5 -.276 -.276 -.5 0 -.5 c .276 -.5 .5 -.276 .5 0 c h}
\def\diaPDF#1#2{#2 0 m #1 #2 l #2 #1 l 0 #2 l h}
\def\credCOLOR   {.54 .14 0}
\def\cblueCOLOR  {.06 .3 .54}
\def\cgreenCOLOR {0 .54 0}
\def\genbox#1#2#3#4#5#6{
    \leavevmode\raise#4bp\hbox to#5bp{\vrule height#5bp depth0bp width0bp
    \pdfliteral{q .5 w \csname #2COLOR\endcsname\space RG
                       \csname #3PDF\endcsname{#5}{#6} S Q
             \ifx1#1 q \csname #2COLOR\endcsname\space rg 
                       \csname #3PDF\endcsname{#5}{#6} f Q\fi}\hss}}

\def\sqbox      #1#2{\genbox{#1}{#2}  {sq}       {0}   {4.5}  {2.25}}
\def\trianbox   #1#2{\genbox{#1}{#2}  {trian}    {0}   {5}    {2.5}}
\def\uptrianbox #1#2{\genbox{#1}{#2}  {uptrian}  {0}   {5}    {2.5}}
\def\circbox    #1#2{\genbox{#1}{#2}  {circ}     {0}   {5}    {2.5}}
\def\diabox     #1#2{\genbox{#1}{#2}  {dia}      {-.5} {6}    {3}}

\begin{abstract}
Graph property prediction tasks are important and numerous. While each task offers a small size of labeled examples, unlabeled graphs have been collected from various sources and at a large scale. A conventional approach is training a model with the unlabeled graphs on self-supervised tasks and then fine-tuning the model on the prediction tasks.
However, the self-supervised task knowledge could not be aligned or sometimes conflicted with what the predictions needed.
In this paper, we propose to extract the knowledge underlying the large set of unlabeled graphs as a specific set of useful data points to augment each property prediction model. We use a diffusion model to fully utilize the unlabeled graphs and design two new objectives to guide the model's denoising process with each task's labeled data to generate task-specific graph examples and their labels. Experiments demonstrate that our data-centric approach performs significantly better than fifteen existing various methods on fifteen tasks. The performance improvement brought by unlabeled data is \textit{visible} as the generated labeled examples unlike the self-supervised learning.

\end{abstract}

\section{Introduction}
\label{sec:introduction}
Graph data such as molecules and polymers are found to have attractive properties in drug and material discovery~\citep{bohm2004scaffold,huang2021therapeutics}, but annotating them requires specialized knowledge, as well as lengthy and costly experiments in wet labs~\citep{cormack2004molecularly}. So, it is important for graph property predictors to learn \textit{useful knowledge} from unlabeled graphs.

Self-supervised learning~\citep{hu2019strategies,rong2020self,you2021graph,kim2022graph} utilizes unlabeled graphs to learn through \textit{predictive tasks} or \textit{contrastive tasks} to represent and transfer the knowledge as \textit{model parameters}. Despite the empirical success in language and vision~\citep{brown2020language,he2022masked}, their performance on graph data applications remains unsatisfactory because of the significant gap between the graph self-supervised task and the graph label prediction task. Models trained on node attribute prediction~\citep{hu2019strategies} as a simple \textit{predictive} self-supervised task extract too limited knowledge from the graph structure, which has been observed after too fast convergence~\citep{sun2022does}.
More complex tasks like graph context prediction~\citep{hu2019strategies,zhang2021motif} may transfer knowledge that conflicts with downstream tasks. Aromatic rings, for instance, are a prevalent structure in molecules~\citep{maziarka2020mol} and are considered valuable in context prediction tasks~\citep{zhang2021motif}. However, graph properties such as oxygen permeability can be more related to non-aromatic rings in some cases~\citep{liu2022graph}, which is overlooked if not using tailored predictive tasks specifically for downstream tasks. As predictive tasks strive for universality, the transferred knowledge may force models to focus more on aromatic rings, leading to poor prediction.

On the other line, \textit{contrastive} tasks~\citep{you2021graph,kim2022graph} aim to learn the similarity between original and perturbed graphs. However, the learned similarity can hardly generalize across tasks~\citep{kim2022graph}. First, perturbations without domain knowledge, \eg bioisosteres, do not preserve broad biological properties~\citep{sun2021mocl}. Second, it is difficult, if not impossible, to find universal perturbations that generalize to diverse property prediction tasks. For example, bioisosteric (subgraph) replacements produce similar biological properties for molecules. And they may reduce toxicity~\citep{brown2014bioisosteres}. So, contrastive tasks with bioisosteric replacement enforce the similarity between toxic and non-toxic molecules. However, models pre-trained on such contrastive tasks hurt the performance on downstream tasks, \eg toxicity prediction.

Our \textit{data-centric} idea avoids the use of self-supervised tasks that are not appropriate. We use a diffusion probabilistic model (known as \textit{diffusion model}) to capture the data distribution of \textit{unlabeled graphs}, leveraging its capability of distribution coverage, stationarity, and scalability~\citep{dhariwal2021diffusion}. At the stage of performing a particular property prediction task, the reverse process, guided by novel task-related optimization objectives, generates new task-specific labeled examples. Minimal sufficient knowledge from the unlabeled data is transferred into these examples, instead of uninterpretable model parameters, and then to enhance the training of prediction models.

\begin{figure}[t]
\captionsetup[subfigure]%
 {labelformat=empty,justification=RaggedRight}
    \begin{subfigure}{0.5\textwidth}
        \begin{minipage}{0.5\linewidth}
        \includegraphics[width=\textwidth]{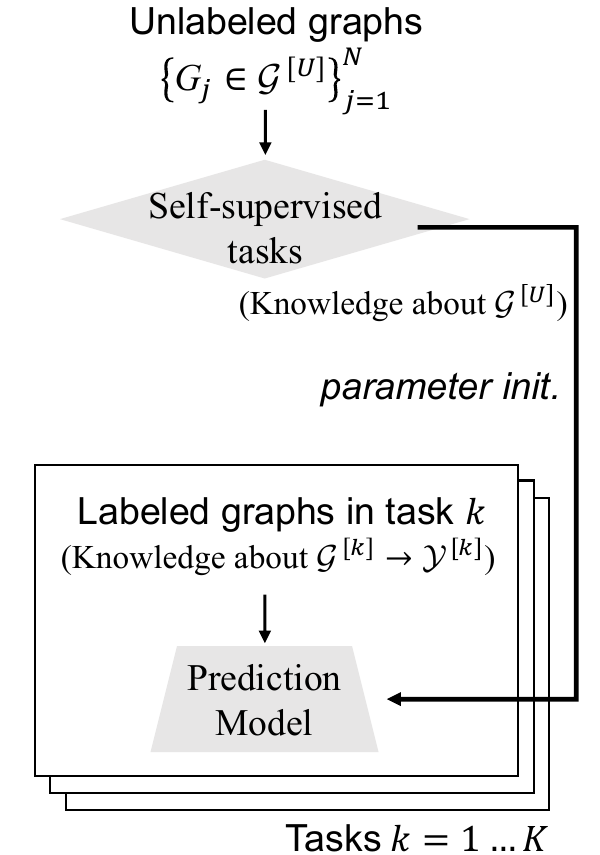}
        \end{minipage}\hfill
        \begin{minipage}{0.5\linewidth}
        \caption{\textbf{Existing approach}: Knowledge from self-supervised tasks could not be aligned or even conflict with what predictions need. Parameter initialization could \textit{hardly interpret} how unlabeled graphs were or would be able to improve the models, leading to high prediction errors.}
        \label{fig:intro_existing}
    \end{minipage}
    \end{subfigure}%
\hfill 
    \begin{subfigure}{0.49\textwidth}
    \begin{minipage}{0.5\linewidth}
    \includegraphics[width=\textwidth]{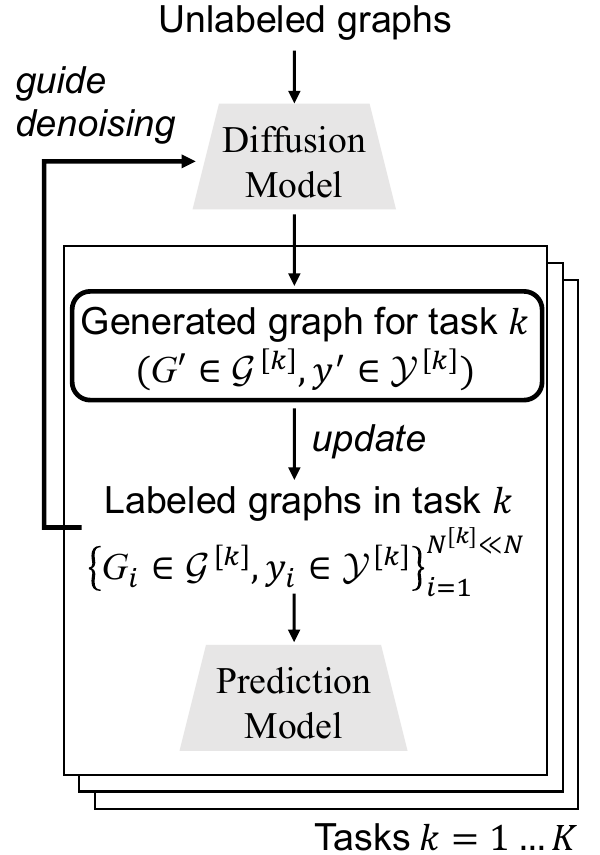}
    \end{minipage}\hfill
    \begin{minipage}{0.5\linewidth}
    \caption{\textbf{Data-centric approach}: Target knowledge in labeled graphs guides the denoising process in diffusion model to generate new labeled examples close to the target graph space instead of the unlabeled graph space. The augmented knowledge for the prediction model is \textit{visible} as graphs.}
    \label{fig:intro_proposed}
    \end{minipage}
\end{subfigure}
\caption{Comparing the diagrams of the existing approach and the proposed approach to learning from unlabeled graphs for a variety of graph property prediction tasks.}
\label{fig:intro}
\end{figure}

To implement our idea, we propose a \textit{Data-Centric Transfer} framework (\method) based on a diffusion model for graph data, as shown in~\cref{fig:intro_proposed}.
It aims to transfer minimal sufficient knowledge from unlabeled graphs to property predictors by data augmentation.
The diffusion model gradually adds Gaussian noise to a graph from which a score function (\ie the gradient of the log probability density) is then learned to estimate the noise step by step to reverse the process. \method trains the diffusion model on the unlabeled graphs to get ready to augment any labeled dataset.
Given a labeled graph from a particular task (\ie type of property), the diffusion model adds noise to perturb it by a few steps and then generates a new graph through the score function.
The new graph could be close to the distribution of the unlabeled graphs for diversity, however, it would lose the relatedness to the target task. So, we add two task-related objectives into the score function to guide the reverse process. When a predictor model $f$ has been trained on the task, given an original labeled graph $G$, the first objective is to optimize the new graph $G^\prime$ to \emph{sufficiently} preserve the label of $G$ with $f$. The second objective is to optimize $G^\prime$ to be very different from (\ie \textit{minimally} similar to) $G$. These two objectives ensure that $G^\prime$ carries minimal sufficient knowledge from the unlabeled graphs to be an augmentation of $G$.
\method iteratively generates new examples to augment the labeled dataset and progressively trains the prediction model with it.

We test \method on \textit{fifteen} graph property prediction datasets from three fields: chemistry (molecules), material science (polymers), and biology (protein-protein interaction graphs). \method achieves the best performance over all these tasks.
We find that the state-of-the-art self-supervised methods often struggle to transfer knowledge to regression tasks, etc. \method reduces the mean absolute error relatively by 13.4\% and 10.2\% compared to the best baseline on the molecule and polymer graph regression tasks, respectively.

\section{Problem Definition}
\label{sec:problem}
Given $K$ property prediction tasks, there are ${N^{[k]}}$ labeled graph examples for the $k$-th task. They are $\{(G_i, y_i) \mid G_i \in \mathcal{G}^{[k]}, y_i \in \mathcal{Y}^{[k]}\}_{i=1}^{N^{[k]}}$, where $\mathcal{G}^{[k]}$  is the graph space and $\mathcal{Y}^{[k]}$ is the label space of the task. 
The prediction model with parameters $\theta$ is defined as $f^{[k]}_\theta: \mathcal{G}^{[k]} \rightarrow \mathcal{Y}^{[k]}$. $f^{[k]}_\theta$ consists of a GNN and a multi-layer perceptron (MLP).
Without the loss of generality, we consider Graph Isomorphism Networks (\gin)~\citep{xu2018powerful} to encode graph structures. Given a graph $G=(\mathcal{V},\mathcal{E}) \in \mathcal{G}^{[k]}$ in the task $k$, \gin updates the representation vector of node $v \in \mathcal{V}$ at $l$-layer:
\begin{equation}\label{eq:gin update msg}
\mathbf{h}_v^{l}=\opfunc{mlp}^{l}\left(\left(1+\epsilon \right) \cdot \mathbf{h}_v^{l-1}+\sum_{u \in \mathcal{N}(v)} \mathbf{h}_u^{l-1}\right),
\end{equation}
where $\epsilon$ is a learnable scalar and $u \in \mathcal{N}(v)$ is one of node $v$'s neighbor nodes.
After stacking $L$ layers, the $\opfunc{readout}$ function (\eg~summation) gets the graph representation across all the nodes. The predicted label is:
\begin{align}
\label{eqn:eq:gin readout and predict}
\begin{split}
\hat{y}&=\opfunc{mlp} \left( \opfunc{readout}\left(\left\{\mathbf{h}^{L}_v \mid v \in G\right\}\right) \right).
\end{split}
\end{align}
$f^{[k]}_\theta$ is hard to be well-trained because it is hard to collect graph labels at a large scale
($N^{[k]}$ is small).

Fortunately, regardless of the tasks, a large number of \textbf{unlabeled graphs} are usually available from the same or similar domains.
Self-supervised learning methods~\citep{hu2019strategies} rely on hand-crafted tasks to extract \textit{knowledge} from the unlabeled examples $\{G_j \in \mathcal{G}^{[U]}, j=1,\dots,N\}$ as \textit{pre-trained model parameters} $\theta$. The uninterpretable parameters are transferred to warm up the prediction models $\{f^{[k]}_\theta\}^{K}_{k=1}$ on the $K$ downstream graph property prediction tasks. 
However, the gap and even conflict between the self-supervised tasks and the property prediction tasks lead to suboptimal performance of the prediction models. In the next section, we present the \method framework that transfers knowledge from the unlabeled graphs with a data-centric approach.

\section{The Data-Centric Transfer Framework}
\label{sec:method}
\subsection{Overview of Proposed Framework}
\label{sec:overview}

The goal of data-centric approaches is to augment training datasets by generating useful labeled data examples. Under that, the goal of our data-centric transfer (\method) framework is to \textit{transfer} the knowledge from unlabeled data into the data augmentation.
Specifically, for each graph-label pair $(G^{[k]} \in \mathcal{G}^{[k]} $, $y^{[k]} \in \mathcal{Y}^{[k]})$ in the task $k$,
the framework is expected to output a new example $G^{\prime[k]}$ with the label $y^{\prime[k]}$ such that (1) $y^{\prime[k]} = y^{[k]}$ and (2) $G^{\prime[k]}$ and $G^{[k]}$ are from the same graph space $\mathcal{G}^{[k]}$. However, if the graph structures of $G^{\prime[k]}$ and $G^{[k]}$ were too similar, the augmentation would duplicate the original data examples, become useless, and even cause over-fitting. So, the optimal graph data augmentation should \textit{enrich the training data with good diversity as well as preserve the labels of the original graphs}.
To achieve this, \method utilizes a diffusion probabilistic model to first \textit{learn the data distribution from unlabeled graphs} (Section~\ref{sec:learning}). Then \method adapts the reverse process in the diffusion model to \textit{generate task-specific labeled graphs for data augmentation} (Section~\ref{sec:generating}).
Thus, the augmented graphs will be derived from the distribution of a huge collection of unlabeled data for \textit{diversity}. To \textit{preserve the labels}, \method controls the reverse process with two {task-related optimization objectives} to transfer \textit{minimal sufficient knowledge} from the unlabeled data. The first objective minimizes an upper bound of mutual information between the augmented and the original graphs in the graph space. The second objective maximizes the probability of the predicted label of augmented graphs being the same as the label of original graphs.
The first is for minimal knowledge transfer, and the second is for sufficient knowledge transfer.
\method integrates the two objectives into the reverse process of the diffusion model to guide the generation of new labeled graphs.
\method iteratively trains the graph property predictor (used in the second objective) and creates the augmented training data. To simplify notations, we remove the task superscript $[k]$ in the following sections.

\def\I{{\mathcal{I}}}
\def\P{{\mathcal{P}}}
\begin{figure*}[t]
    \centering
    \begin{minipage}[c]{0.63\textwidth}
    \includegraphics[width=\textwidth]{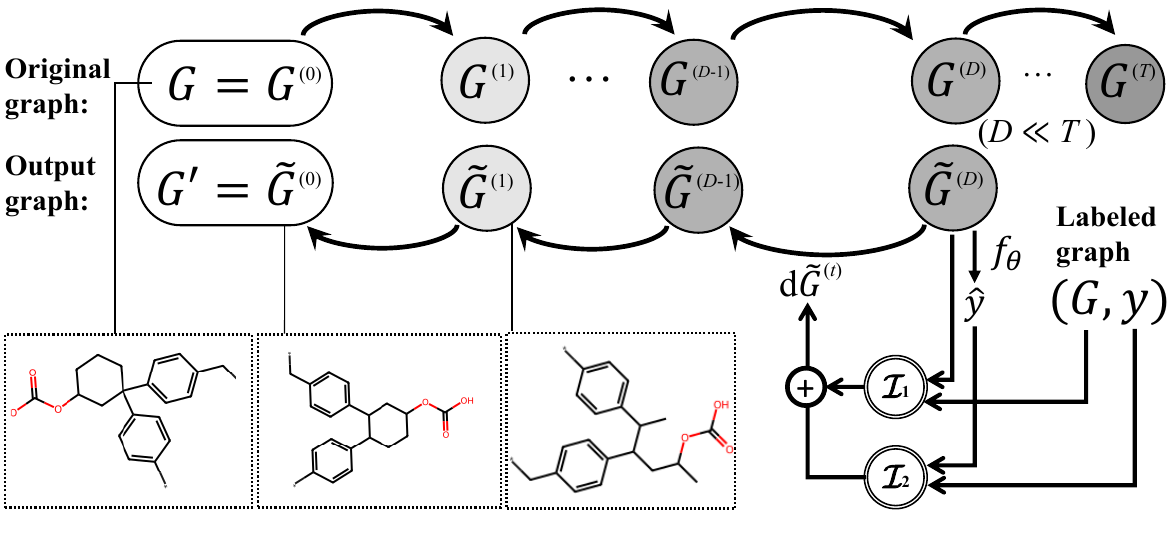}
    \end{minipage}\hfill
    \begin{minipage}[c]{0.36\textwidth}
    \caption{Diffusion model in DCT: It performs task-specific data augmentation using objectives $\I_1$ and $\I_2$ in the reverse process. The model was trained on unlabeled graphs to learn the general data distribution. Then it generates $(G^\prime,y^\prime=y)$ based on $(G,y)$ in the reverse process. It perturbs $G$ with $D$ steps and optimizes $G^\prime$ to be minimally similar to $G$ (Objective $\I_1$) and sufficiently preserve the label of $G$ (Objective $\I_2$).
    } \label{fig:implementation}
  \end{minipage}
\end{figure*}
\begin{figure*}[t]
    \centering
    \includegraphics[width=0.99\linewidth]{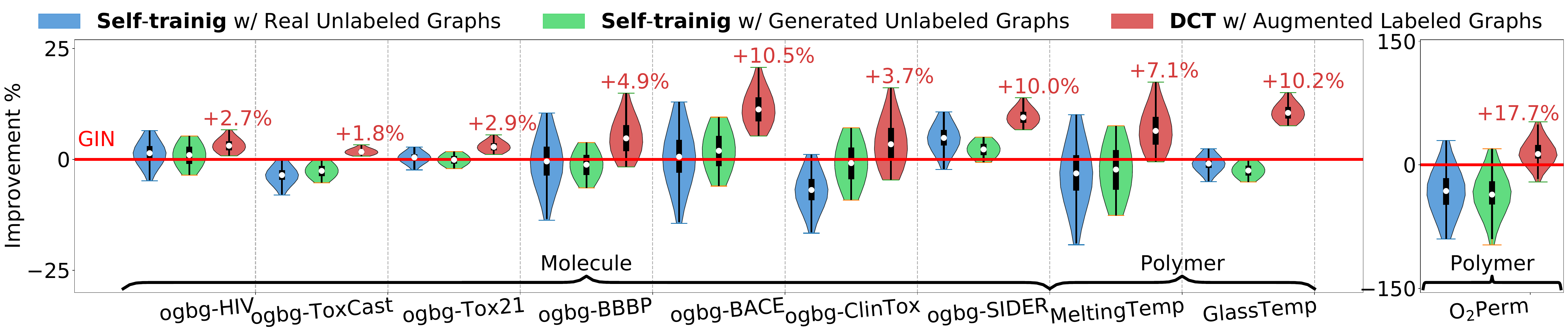}
    \caption{
    Relative improvement (increased AUC or reduced MAE) from three data-centric methods (over ten runs), compared to the basic GIN: Blue is for self-training with selected real unlabeled graphs. Green is for self-training with graphs directly generated by a standard diffusion model. Red is for \method that generates task-specific labeled graphs. The first two often make little or negative impact. Our \method has consistent and significant improvement shown as the percentages in \textcolor[rgb]{0.83137255, 0.22745098, 0.22745098}{red}.
    }
    \label{fig:compare gnn runs}
\end{figure*}
\subsection{Learning Data Distribution from Unlabeled Graphs}
\label{sec:learning}

The diffusion process for graphs in~\cref{fig:implementation} applies to both graph structure and node features. The diffusion model slowly corrupts unlabeled graphs to a standard normal distribution with noise. For graph generation, the model samples noise from the normal distribution and learns a score function to reverse the perturbed noise. Given an unlabeled graph $G$, we use continuous time $t \in [0,T]$ to index multiple diffusion steps $\{G^{(t)}\}_{t=1}^T$ on the graph, such that $G^{(0)}$ follows the original data distribution and $G^{(T)}$ follows a prior distribution like the normal distribution. The forward diffusion is a stochastic differential equation (SDE) from the graph to the noise:
\begin{equation}\label{eq:forward graph SDE}
\mathrm{d} G^{(t)}=\mathbf{f} \left(G^{(t)}, t \right) \mathrm{d} t+g(t)~\mathrm{d} \mathbf{w},
\end{equation}
where $\mathbf{w}$ is the standard Wiener process, $\mathbf{f}(\cdot, t): \mathcal{G} \rightarrow \mathcal{G}$ is the drift coefficient and $g(t): \mathbb{R} \rightarrow \mathbb{R}$ is the diffusion coefficient. $\mathbf{f}(G^{(t)}, t)$ and $g(t)$ relate to the amount of noise added to the graph at each infinitesimal step $t$. The reverse-time SDE uses gradient fields or scores of the perturbed graphs $\nabla_{G^{(t)}} \log p_t(G^{(t)})$ for denoising and graph generation from $T$ to $0$~\citep{song2020score}:
\begin{equation}\label{eq:reverse graph SDE}
\mathrm{d} G^{(t)} =\left[\mathbf{f}(G^{(t)}, t)-g(t)^2 \nabla_{G^{(t)}} \log p_t(G^{(t)})\right] \mathrm{d} t+g(t) \mathrm{d} \overline{\mathbf{w}},
\end{equation}
where $p_t(G^{(t)})$ is the marginal distribution at time $t$ in forward diffusion. $\overline{\mathbf{w}}$ is a reverse time standard Wiener process. $\mathrm{d} t$ here is an infinitesimal negative time step. The score $\nabla_{G^{(t)}} \log p_t(G^{(t)})$ is unknown in practice and it is approximated by the score function $\mathbf{s}(G^{(t)}, t)$ with score matching techniques~\citep{song2020score}. On graphs, \citet{jo2022score} used two GNNs to develop the score function $\mathbf{s}(G^{(t)}, t)$ to de-noise both node features and graph structures and details are in~\cref{add:method:tech}.

\subsection{Generating Task-specific Labeled Graphs}
\label{sec:generating}

\vspace{-0.05in}
\textit{Self-training} approaches would propose to either (1) select unlabeled graphs by a graph property predictor or (2) generate graphs directly from the standard diffusion model, and then use the predictor to assign them labels so that the training data could be enriched.
However, we have observed that neither of them can guarantee positive impact on the prediction performance. In fact, as shown in~\cref{fig:compare gnn runs}, they make very little or even negative impact.
That is because \textit{the selected or directly-generated graphs are too different from the labeled graph space of the target tasks}. Task details of ten datasets are in~\cref{sec: add exp setups}.

Given a labeled graph ($G,y$) from the original dataset of a specific task, the new labeled graph $(G^\prime,y^\prime)$ is expected to provide \textit{useful knowledge to augment} the training set. We name it \emph{the augmented graph} throughout this section. The augmented graph is desired to have the following two properties, as in~\cref{sec:overview}: \textbf{Task relatedness}: As an effective training data point, $G^\prime \in \mathcal{G}$ and $y^\prime \in \mathcal{Y}$ are from the graph/label spaces of the specific task where $(G,y)$ come from and thus transfer sufficient task knowledge into the training set; \textbf{Diversity}: If $G^\prime$ was too similar to $G$, the new data point would cause severe over-fitting on the property prediction model. The augmentation aims to learn from unlabeled graph to create diverse data points, which should contain minimal task knowledge about $G$.

The selected unlabeled graphs used in \textit{self-training} have little task relatedness because the unlabeled data distribution might be too far from the one of the specific task. Existing graph \textit{data augmentation} methods could not create diverse graph examples because they manipulated labeled graphs and did not learn from the unlabeled graphs. Our novel data-centric approach \method works towards both desired properties by transferring \textit{minimally sufficient knowledge} from the unlabeled graphs: \textbf{Sufficiency} is achieved by maximizing the possibility for label preservation (i.e., $y^\prime = y$). It ensures that the knowledge from unlabeled graphs is task-related;
\textbf{Minimality} refers to the minimization of graph similarity between $G^\prime$ and $G$ to ensure that the augmentation introduces diversity. Both optimizations can be formulated using mutual information $\I(\cdot \ ; \cdot)$ to generate task-specific labeled data $(G^\prime,y^\prime)$:
\begin{definition}[Sufficiency for Data Augmentation]
    The augmented graph $G^\prime$ sufficiently preserves the label of the original graph $G$ if and only if $\I(G^\prime;y) = \I(G;y) $.
\end{definition}

\begin{definition}[Minimal Sufficiency for Data Augmentation]
    The Sufficiency is minimal for data augmentation if and only if $\I(G^\prime; G) \leq \I(\bar{G}; G)$, $\forall \bar{G}$ represents any augmented graph that sufficiently preserves the original graph's label.
\end{definition}

Self-supervised tasks applied a similar philosophy in pre-training~\citep{soatto2014visual}, however, they did not use labeled data from any specific tasks. So the optimizations were unable to extract useful knowledge and transfer it to the downstream~\citep{tian2020makes}. In our \method that performs task-specific data augmentation, the augmented graphs can be optimized toward the objectives using any labeled graph $G$ and its label $y$:
\begin{equation}\label{eq:fine-tune objective}
    \min_{\I_1}\max_{\I_2} \ \ \mathbb{E}_{G} \left[ 
    \I_1\left(G^\prime; G \right) + \I_2\left(G^\prime; y\right) 
    \right].
\end{equation}
For the first objective, we use the leave-one-out variant of {InfoNCE}~\citep{poole2019variational,oord2018representation} as the upper bound estimation. For the $i$-th labeled graph $(G_i,y_i)$,
\begin{equation}\label{eq:upper bound infonce}
    \I_1 \leq \I_\text{bound} (G_i^\prime; G_i) = 
    \log \frac{p(G_i^\prime |G_i)}{\sum_{j=1,j \neq i}^{M} p(G_i^\prime | G_j)} ,
\end{equation}
where
$G_i^\prime$ is the augmented graph.
When $G_i^\prime$ is optimized, $G_i$ makes a positive pair; $\{G_j\}$ ($j\neq i$) are $M-1$ negative samples of labels that do not equal $y_i$. ($M$ is a hyperparameter.)
We use cosine similarity and a softmax function to calculate $p(G_i^\prime|G_j)=\frac{\exp({\operatorname{sim}(G_i^\prime, G_j) })}{\sum_{j=1}^M \exp({\operatorname{sim}(G_i^\prime, G_j) }) }$.  
In practice, we extract statistical features of graphs to calculate their similarity. Details are in \cref{add:sec:raw feature}.

For the second objective, we denote the predicted label of the augmented graph $G^\prime$ by $f_\theta(G^\prime)$.
We maximize the log likelihood $\operatorname{log} p \left(y | f_\theta(G^\prime) \right)$ to maximize $\I_2(G^\prime; y)$.
Specifically, after the predictor $f_\theta$ is trained for several epochs on the labeled data, we freeze its parameters and use it to optimize the augmented graphs so they are task-related:
\begin{equation}\label{eq:fine-tune loss}
    \mathcal{L} (G^\prime) = 
    \I_\text{bound}\left(G^\prime; G \right) 
    - \operatorname{log} p \left(y | f_\theta(G^\prime) \right).
\end{equation}

\paragraph{Framework details:} As shown in~\cref{fig:implementation}, after the diffusion model learns the data distribution from unlabeled graphs, given a labeled graph $G$ from a specific task, \method perturbs it for $D$ ($D \ll T$) steps. The perturbed noisy graph, denoted by $\Tilde{G}^{(D)}$, stays inside the task-specific graph and label space, rather than the noise distribution (at step $T$). To reverse the noise in it and generate a task-specific augmented example $G^\prime$, \method integrates the loss function in~\cref{eq:fine-tune loss} into the score function $\mathbf{s}(\cdot, t)$ for minimal sufficient knowledge transfer:
\begin{align}\label{eq:finetune guided reverse SDE}
    \mathrm{d} \Tilde{G}^{(t)} = \left[\mathbf{f}(\Tilde{G}^{(t)}, t) - g(t)^2 \left(\mathbf{s}(\Tilde{G}^{(t)}, t) - \alpha \nabla_{\Tilde{G}^{(t)}} \mathcal{L}(\Tilde{G}^{(t)}) \right) \right] \mathrm{d}t
    +g(t) \mathrm{d} \overline{\mathbf{w}},
\end{align}
where $\alpha$ is a scalar for score alignment between $\mathbf{s}$ and $\nabla\mathcal{L}$ to avoid the dominance of any of them: $\alpha = \frac{\| \mathbf{s}(\Tilde{G}^{(t)}, t) \|_2}
    { \| \nabla_{\Tilde{G}^{(t)}} \mathcal{L}(\Tilde{G}^{(t)}) \|_2 }.$
Because $\Tilde{G}^{(t)}$ is an intermediate state in the reverse process, the noise in it may fail the optimizations. So, we design a new sampling method named \textit{double-loop sampling} for accurate loss calculation. 
It has an inner-loop sampling using \cref{eq:reverse graph SDE} to sample $\hat{G}_{(t)}$, as the denoised version of $\Tilde{G}^{(t)}$ at the reverse time $t$. Then $\nabla_{\hat{G}} \mathcal{L}( \hat{G}_{(t)} )$ is calculated as an alternative for $ \nabla_{\Tilde{G}^{(t)}} \mathcal{L}(\Tilde{G}^{ (t)})$.
Finally, an outer-loop sampling takes one step to guide denoising using~\cref{eq:finetune guided reverse SDE}.

DCT iteratively creates the augmented graphs $(G^\prime,y^\prime)$, updates the training dataset $\{(G_i, y_i)\}$, and trains the graph property predictor $f_\theta$. In each iteration, for task $k$, $n \ll N^{[k]}$ labeled graphs of the lowest property prediction loss are selected to create the augmented graphs.The predictor is better fitted to these graphs for more accurate sufficiency estimation of the augmentation.

\vspace{-0.04in}
\section{Experiments}
\label{sec:experiments}
\begin{table}[t]
\caption{Statistics of datasets for graph property prediction in different domains.}
\label{tab:dataset_stat}
    \centering
    \begin{adjustbox}{width=0.9\linewidth}
    \begin{tabular}{lllllcc}
    \toprule
       Data Type &  Dataset & \# Graphs & Prediction Task & \# Task & Avg./Max \# Nodes & Avg./Max \# Edges \\
        \midrule
        \multirow{10}{*}{{Molecules}} & 
        \hiv & 41,127 & Classification & 1 & 25.5 / 222 & 54.9 / 502 \\
        & \toxcast & 8,576 & Classification & 617 & 18.8 / 124 & 38.5 / 268 \\
        & \toxt & 7,831 & Classification & 12 & 18.6 / 132 & 38.6 / 290 \\
        & \bbbp & 2,039 & Classification & 1 & 24.1 / 132 & 51.9 / 290 \\
        & \bace & 1,513 & Classification & 1 & 34.1 / 97 & 73.7 / 202 \\
        & \clintox & 1,477 & Classification & 2 & 26.2 / 136 & 55.8 / 286 \\
        & \sider & 1,427 & Classification & 27 & 33.6 / 492 & 70.7 / 1010 \\
        & \lipo & 4200 & Regression & 1 & 27 / 115 & 59 / 236 \\
        & \esol & 1128  & Regression & 1 & 13.3 / 55 & 27.4 / 124 \\ 
        & \freesolv & 642 & Regression & 1 &  8.7 / 24 & 16.8 / 50 \\
        \midrule
        \multirow{4}{*}{{Polymers}} 
        &  \glass & 7,174 & Regression & 1 & 36.7 / 166 & 79.3 / 362 \\
        &  \melting & 3,651 & Regression & 1 & 26.9 / 102 & 55.4 / 212 \\
        & \thermal & 759 & Regression & 1 & 21.3 / 71 & 42.3 / 162 \\
        & \oxygen & 595 & Regression & 1 & 37.3 / 103 & 82.1 / 234 \\
        \midrule
        Proteins & \ppi & 88000 & Classification & 40 & 49.4 / 111 & 890.8 / 11556 \\        
    \bottomrule
    \end{tabular}
    \end{adjustbox}
    \vspace{-0.2in}
\end{table}
In this section, we present and analyze experimental results to demonstrate the outstanding performance of \method, the usefulness of new optimization objectives, the effect of hyperparameters and iterative process, and the interpretability of ``visible'' knowledge transfer from unlabeled graphs.

\subsection{Experimental Setup}
\paragraph{Tasks and metrics:} 
Experiments are conducted on 15 graph property prediction tasks in chemistry, material science, and biology, including seven molecule classification, three molecule regression tasks from open graph benchmarks~\citep{hu2020open}, four polymer regression tasks, and protein function prediction (\ppi)~\citep{hu2019strategies}. Dataset statistics is presented in~\cref{tab:dataset_stat}. We use the area under the ROC curve (AUC) to evaluate classifiers and mean absolute error (MAE) for regressors. 

\paragraph{Baselines and implementation:}
Besides \gin, there are three lines of baseline methods: (1) \textit{self-supervised learning methods} including \edgepred, \attrmask, \contextpred in~\citep{hu2019strategies}, \infomax~\citep{velickovic2019deep}, \joao~\citep{you2021graph}, \graphlog~\citep{xu2021self}, \mgssl~\cite{zhang2021motif} and \dsla~\citep{kim2022graph}, (2) \textit{semi-supervised learning methods} including self-training with selected unlabeled graphs (\streal) and generated graphs (\stgen) and \infograph~\citep{sun2019infograph}, and (3) \textit{graph data augmentation (GDA) methods} including \flag~\citep{kong2022robust}, \grea~\citep{liu2022graph}, and \gmix~\citep{han2022g}.
For self-supervised pre-training, we follow their own settings and directly use their pre-trained models if available. For semi-supervised learning methods and \method, we use 113K QM9~\citep{ramakrishnan2014quantum} and 306K \ppi graphs~\citep{hu2019strategies} as unlabeled data sources for the tasks on molecules/polymers and proteins, respectively. For \method, we tune three major hyper-parameters: the number of perturbation steps $D \in [1,10]$, the number of negative samples $M \in [1,10]$, and top-$n$~\% labeled graphs of lowest property prediction loss selected for data augmentation.

\definecolor{LightCyan}{rgb}{0.88,1,1}
\definecolor{negative}{gray}{0.92}
\begin{table*}[t!]
    \renewcommand{\arraystretch}{0.95}
    \renewcommand{\tabcolsep}{1.2mm}
    \caption{\small {Mean\scriptsize(Std)} on tasks from different fields. The best mean is \textbf{bold}. The best baseline is \underline{underlined}. Results are \colorbox{negative}{highlighted} if unlabeled graphs bring significant negative impacts compared to \gin. The MAE for \thermal is scaled $\times$ 100. \gmix was proposed for classification. \mgssl was proposed for molecules.}
    \centering
    \vspace{-0.1in}
    \begin{adjustbox}{width=0.95\textwidth}
    \begin{tabular}{llccccccc}
    \toprule
    \multicolumn{2}{c}{} & \multicolumn{7}{c}{Molecule Classification: AUC (\%) $\uparrow$} \\
    \cmidrule{3-9}
    &  & \hiv & \toxcast & \toxt & \bbbp & \bace & \clintox & \sider \\
    \multicolumn{2}{c}{\# Training Graphs} & 32,901 & 6,860 & 6,264 & 1,631 & 1,210 & 1,181 & 1,141 \\
    \midrule
    & \gin & 77.4{\scriptsize (1.2)} & 66.9{\scriptsize (0.2)} & 76.0{\scriptsize (0.6)} & 67.5{\scriptsize (2.7)} & 77.5{\scriptsize (2.8)} & \underline{88.8}{\scriptsize (3.8)} & 58.1{\scriptsize (0.9)} \\
    \midrule
    
    \multirow{7}{*}{\rotatebox{90}{\textit{\fontsize{8pt}{8pt} Self-Supervised}}}

    & \edgepred 
    & 78.1{\scriptsize (1.3)} & \cellcolor{negative} 63.9{\scriptsize (0.4)} & 75.5{\scriptsize (0.4)} & 69.9{\scriptsize (0.5)} & 79.5{\scriptsize (1.0)} & \cellcolor{negative} 62.9{\scriptsize (2.3)} & 59.7{\scriptsize (0.8)} \\
    & \attrmask 
    & 77.1{\scriptsize (1.7)} & \cellcolor{negative} 64.2{\scriptsize (0.5)} & 76.6{\scriptsize (0.4)} & \cellcolor{negative} 63.9{\scriptsize (1.2)} & 79.3{\scriptsize (0.7)} & \cellcolor{negative} 70.4{\scriptsize (1.1)} & 60.7{\scriptsize (0.4)} \\
    & \contextpred 
    & 78.4{\scriptsize (0.1)} & \cellcolor{negative} 63.7{\scriptsize (0.3)} & \cellcolor{negative} 75.0{\scriptsize (0.1)} & 68.8{\scriptsize (1.6)} & \cellcolor{negative} 75.7{\scriptsize (1.0)} & \cellcolor{negative} 63.2{\scriptsize (6.5)} & 60.7{\scriptsize (0.8)} \\
    & \infomax 
    & \cellcolor{negative} 75.4{\scriptsize (1.8)} & \cellcolor{negative} 61.7{\scriptsize (1.0)} & \cellcolor{negative} 75.5{\scriptsize (0.4)} & 69.2{\scriptsize (0.5)} & \cellcolor{negative} 76.8{\scriptsize (0.2)} & \cellcolor{negative} 73.0{\scriptsize (0.2)} & 58.6{\scriptsize (0.5)} \\
    & \joao 
    & \cellcolor{negative} 76.2{\scriptsize (0.2)} & \cellcolor{negative} 64.8{\scriptsize (0.3)} & \cellcolor{negative} 74.8{\scriptsize (0.5)} & 69.3{\scriptsize (2.5)} & \cellcolor{negative} 75.9{\scriptsize (3.9)} & \cellcolor{negative} 69.4{\scriptsize (4.5)} & 60.8{\scriptsize (0.6)}\\
    & \graphlog 
    & \cellcolor{negative} 74.8{\scriptsize (1.1)} & \cellcolor{negative} 63.2{\scriptsize (0.8)} & 75.4{\scriptsize (0.8)} & 67.5{\scriptsize (2.3)} & 80.4{\scriptsize (3.6)} & \cellcolor{negative} 69.0{\scriptsize (6.6)}& \cellcolor{negative} 57.0{\scriptsize (0.9)} \\
    & \mgssl 
    & 77.1{\scriptsize (1.1)} & \cellcolor{negative} 65.7{\scriptsize (0.4)} & 77.2{\scriptsize (0.3)} & 66.9{\scriptsize (0.9)} & 81.3{\scriptsize (2.4)} & \cellcolor{negative} 69.8{\scriptsize (5.0)} & \underline{63.6}{\scriptsize (1.0)} \\
    
    & \dsla 
    & \cellcolor{negative} 76.9{\scriptsize (0.9)} & \cellcolor{negative} 60.8{\scriptsize (1.2)} & 76.1{\scriptsize (0.1)} & \cellcolor{negative} 62.6{\scriptsize (1.0)} & 80.3{\scriptsize (0.6)} & \cellcolor{negative} 78.3{\scriptsize (2.4)} & \cellcolor{negative} 55.1{\scriptsize (1.0)} \\
    
    \midrule
    \multirow{3}{*}{\rotatebox{90}{\textit{\fontsize{8pt}{8pt} Semi-SL}}}
    
    & \infograph 
    & \cellcolor{negative} 73.3{\scriptsize (0.7)} & \cellcolor{negative} 61.5{\scriptsize (1.1)} & \cellcolor{negative} 67.6{\scriptsize (0.9)} & \cellcolor{negative} 61.6{\scriptsize (4.4)} & \cellcolor{negative} 75.9{\scriptsize (1.8)} & \cellcolor{negative} 62.2{\scriptsize (5.5)} & \cellcolor{negative} 56.3{\scriptsize (2.3)} \\ 

    & \streal
    & 78.3{\scriptsize (0.6)} & \cellcolor{negative} 64.5{\scriptsize (1.0)} & 76.2{\scriptsize (0.5)} & 66.7{\scriptsize (1.9)} & 77.4{\scriptsize (1.8)} & \cellcolor{negative} 82.2{\scriptsize (2.4)} & 60.8{\scriptsize (1.2)} \\
    & \stgen 
    & 77.9{\scriptsize (1.6)} & \cellcolor{negative} 65.1{\scriptsize (1.0)} & 75.8{\scriptsize (0.9)} & \cellcolor{negative} 66.3{\scriptsize (1.5)} & 78.4{\scriptsize (3.0)} & \cellcolor{negative} 87.3{\scriptsize (1.3)} & 59.3{\scriptsize (1.3)} \\

    \midrule
    \multirow{3}{*}{\rotatebox{90}{\textit{\fontsize{8pt}{8pt} GDA }}} 
   
    & \flag 
    & 74.6{\scriptsize (1.7)} & 59.9{\scriptsize (1.6)} & 76.9{\scriptsize (0.7)} & 66.6{\scriptsize (1.0)} & 79.1{\scriptsize (1.2)} & 85.1{\scriptsize (3.4)} & 57.6{\scriptsize (2.3)} \\
    & \grea 
    & \underline{79.3}{\scriptsize (0.9)} & \underline{67.5}{\scriptsize (0.7)} & \underline{77.2}{\scriptsize (1.2)} & 69.7{\scriptsize (1.3)} & \underline{82.4}{\scriptsize (2.4)} & 87.9{\scriptsize (3.7)} & 60.1{\scriptsize (2.0)} \\
    & \gmix 
    & 77.1{\scriptsize (1.1)} & 55.6{\scriptsize (1.1)} & 64.6{\scriptsize (0.4)} & \underline{70.2}{\scriptsize (1.0)}  & 77.8{\scriptsize (3.3)} & 60.2{\scriptsize (7.5)} & 56.8{\scriptsize (3.5)} \\
    
    \cmidrule{1-9}
    & \method~(Ours) 
    & \textbf{79.5}{\scriptsize (1.0)} & \textbf{68.1}{\scriptsize (0.2)} & \textbf{78.2}{\scriptsize (0.2)} & \textbf{70.8}{\scriptsize (0.5)} & \textbf{85.6}{\scriptsize (0.6)} & \textbf{92.1}{\scriptsize (0.8)} & \textbf{63.9}{\scriptsize (0.3)} \\
    \end{tabular}
    \end{adjustbox}
    
    \begin{adjustbox}{width=1\textwidth}
    \begin{tabular}{llcccccccc}
    \toprule
    \multicolumn{2}{c}{} & \multicolumn{3}{c}{{Molecule Regression: MAE $\downarrow$}} & \multicolumn{4}{c}{{Polymer Regression: MAE $\downarrow$}} & Bio: AUC {(\%)}$ \uparrow$ \\
    \cmidrule[0.4pt](lr){3-5} \cmidrule[0.4pt](lr){6-9} \cmidrule[0.4pt](lr){10-10}
    & & \lipo & \esol & \freesolv & \glass & \melting & \thermal & \oxygen & \ppi \\
    \multicolumn{2}{c}{\# Training Graphs} & 3,360 & 902 & 513 & 4,303 & 2,189 & 455 & 356 & 60,715 \\
    \midrule
    
    & \gin & 0.545{\scriptsize (0.019)} & 0.766{\scriptsize (0.016)} & 1.639{\scriptsize (0.146)} 
    & \underline{26.4}{\scriptsize (0.2)} & \underline{40.9}{\scriptsize (2.2)} & 3.25{\scriptsize (0.19)} & 201.3{\scriptsize (45.0)} & 
    {69.1}{\scriptsize (0.0)}
    \\
    
    \midrule
    \multirow{7}{*}{\rotatebox{90}{\textit{\fontsize{8pt}{8pt} Self-Supervised}}}
    
    & \edgepred 
    & \cellcolor{negative} 0.585{\scriptsize (0.008)} & \cellcolor{negative} 1.062{\scriptsize (0.066)} & \cellcolor{negative} 2.249{\scriptsize (0.150)} 
    & \cellcolor{negative} 27.6{\scriptsize (1.4)} & \cellcolor{negative} 47.4{\scriptsize (2.8)} & \cellcolor{negative} 3.69{\scriptsize (0.50)} & 207.3{\scriptsize (41.7)} & \cellcolor{negative} 63.7{\scriptsize (1.1)} \\
    & \attrmask 
    & \cellcolor{negative} 0.573{\scriptsize (0.009)} & \cellcolor{negative} 1.041{\scriptsize (0.041)} & \cellcolor{negative} 1.952{\scriptsize (0.088)}
    & \cellcolor{negative}  27.7{\scriptsize (0.8)} & \cellcolor{negative} 45.8{\scriptsize (2.6)} & 3.17{\scriptsize (0.32)} & {179.9}{\scriptsize (30.8)} & \cellcolor{negative} 64.1{\scriptsize (1.8)} \\
    & \contextpred 
    & \cellcolor{negative} 0.592{\scriptsize (0.007)} & \cellcolor{negative} 0.971{\scriptsize (0.027)} & \cellcolor{negative} 2.193{\scriptsize (0.151)} 
    & \cellcolor{negative} 27.6{\scriptsize (0.3)} & \cellcolor{negative} 46.7{\scriptsize (1.9)} & 3.15{\scriptsize (0.24)} & 191.2{\scriptsize (35.2)} & \cellcolor{negative} 62.0{\scriptsize (1.2)} \\
    & \infomax 
    & \cellcolor{negative} 0.581{\scriptsize (0.009)} & \cellcolor{negative} 0.935{\scriptsize (0.018)} & \cellcolor{negative} 2.197{\scriptsize (0.129)} 
    & \cellcolor{negative}  27.5{\scriptsize (0.8)} & \cellcolor{negative} 46.5{\scriptsize (2.8)} & 3.31{\scriptsize (0.25)} & \cellcolor{negative} 231.0{\scriptsize (52.6)} & \cellcolor{negative} 63.3{\scriptsize (1.2)} \\
    & \joao 
    & \cellcolor{negative} 0.596{\scriptsize (0.016)} & \cellcolor{negative} 1.098{\scriptsize (0.037)} & \cellcolor{negative} 2.465{\scriptsize (0.095)} 
    & \cellcolor{negative}  27.5{\scriptsize (0.2)}& \cellcolor{negative} 46.0{\scriptsize (0.2)} & \cellcolor{negative} 3.55{\scriptsize (0.26)} & 207.7{\scriptsize (43.7)} & \cellcolor{negative} 61.5{\scriptsize (1.2)} \\
    & \graphlog 
    & \cellcolor{negative} 0.577{\scriptsize (0.010)} & \cellcolor{negative} 1.109{\scriptsize (0.059)} & \cellcolor{negative} 2.373{\scriptsize (0.283)} 
    & \cellcolor{negative}  29.5{\scriptsize (1.3)} & \cellcolor{negative} 50.3{\scriptsize (3.3)} & 3.01{\scriptsize (0.17)} & \cellcolor{negative} 229.7{\scriptsize (48.3)} & \cellcolor{negative} 62.1{\scriptsize (0.6)} \\
    & \mgssl 
    & \cellcolor{negative} 0.569{\scriptsize (0.007)} & \cellcolor{negative} 0.998{\scriptsize (0.031)} & \cellcolor{negative} 1.956{\scriptsize (0.077)} & \cellcolor{negative} 26.9{\scriptsize (0.4)} & \cellcolor{negative} 42.7{\scriptsize (1.2)} & 3.10{\scriptsize (0.14)} & 201.1{\scriptsize (31.9)} & N.A. \\
    
    & \dsla 
    & \cellcolor{negative} 0.563{\scriptsize (0.004)} & \cellcolor{negative} 1.064{\scriptsize (0.030)} & \cellcolor{negative} 2.190{\scriptsize (0.149)} 
    & \cellcolor{negative}  27.5{\scriptsize (1.0)} & \cellcolor{negative} 51.7{\scriptsize (2.5)}  & 2.71{\scriptsize (0.08)} & \cellcolor{negative} 257.8{\scriptsize (30.2)} & \cellcolor{negative} 65.0{\scriptsize (1.2)} \\

    \midrule
    \multirow{3}{*}{\rotatebox{90}{\textit{\fontsize{8pt}{8pt} Semi-SL}}}
    & \infograph 
    & \cellcolor{negative} 0.793{\scriptsize (0.094)} & \cellcolor{negative} 1.285{\scriptsize (0.093)} & \cellcolor{negative} 3.710{\scriptsize (0.418)}  
    & \cellcolor{negative}  30.8{\scriptsize (1.2)} & \cellcolor{negative} 51.2{\scriptsize (5.1)} & 2.75{\scriptsize (0.15)} & 207.2{\scriptsize (21.8)} & \cellcolor{negative} 67.7{\scriptsize (0.4)} \\
    
    & \streal 
    & \underline{0.526}{\scriptsize (0.009)} & \cellcolor{negative} 0.788{\scriptsize (0.070)} & 1.770{\scriptsize (0.251)}
    & 26.6{\scriptsize (0.3)} & 42.3{\scriptsize (1.2)} & \underline{2.64}{\scriptsize (0.07)} & \cellcolor{negative} 256.0{\scriptsize (17.5)} & 68.9{\scriptsize (0.1)} \\
    & \stgen &
    0.531{\scriptsize (0.031)} & \underline{0.724}{\scriptsize (0.082)} & \underline{1.547}{\scriptsize (0.082)} 
    & 26.8{\scriptsize (0.3)} & 42.0{\scriptsize (0.9)} & 2.70{\scriptsize (0.03)}& \cellcolor{negative} 262.2{\scriptsize (10.1)} & 68.6{\scriptsize (0.6)} \\
    \midrule
    \multirow{2}{*}{\rotatebox{90}{\textit{\fontsize{8pt}{8pt} GDA}}} 
    
    & \flag 
    & 0.528{\scriptsize (0.012)} & 0.755{\scriptsize (0.039)} & 1.565{\scriptsize (0.098)} 
    & 26.6{\scriptsize (1.3)}  & 44.2{\scriptsize (2.0)} & 3.05{\scriptsize (0.10)} & \underline{177.7}{\scriptsize (60.7)} & \underline{69.2}{\scriptsize (0.2)} \\
    & \grea 
    & 0.586{\scriptsize (0.036)} & 0.805{\scriptsize (0.135)} & 1.829{\scriptsize (0.368)} 
    & 26.7{\scriptsize (1.0)}  & 41.1{\scriptsize (0.8)} & 3.23{\scriptsize (0.18)}  & 194.0{\scriptsize (45.5)} & 68.8{\scriptsize (0.2)} \\

    \cmidrule{1-10}
    & \method~(Ours) 
    & \textbf{0.516}{\scriptsize (0.071)} & \textbf{0.717}{\scriptsize (0.020)} & \textbf{1.339}{\scriptsize (0.075)}
    & \textbf{23.7}{\scriptsize (0.2)} & \textbf{38.0}{\scriptsize (0.8)} & \textbf{2.59}{\scriptsize (0.11)} & \textbf{165.6}{\scriptsize (24.3)} & \textbf{69.5}{\scriptsize (0.2)} \\
    \bottomrule
    \end{tabular}
    \end{adjustbox}
    \vspace{-0.2in}
    \label{tab:result of molecules}
\end{table*}

\subsection{Outstanding Property Prediction Performance}
We report the model performance using mean and standard deviation over 10 runs~\cref{tab:result of molecules}. \method is the best on all 15 tasks compared to the state-of-the-art baselines. Our observations are:

\textbf{(1) \gin is the most competitive baseline and outperforms self-supervised learning methods.} On 7 of 15 tasks, \gin outperforms all the 7 self-supervised learning methods. Because self-supervised pre-training imposes constraints on the model architecture, it undermines the true power of GNNs and under-performs the GNNs that are properly used. 

\textbf{(2) Self-training and GDA methods perform better than \gin but cannot effectively learn from unlabeled data.} Self-training (\streal and \stgen) is often the best baseline in regression tasks.
GDA (\grea and \gmix) methods outperform self-training in most classification tasks except \sider,
because they are often designed to exploit categorical labeled data and remain under-explored for regression.
Although self-training benefits from selecting unlabeled examples in some graph regression tasks, they are \textit{negatively} affected by the unlabeled graphs in the classification tasks such as \toxcast and \clintox.
As indicated in~\cref{fig:compare gnn runs}, it is inappropriate to pseudo-label unlabeled graphs in self-training due to the huge gap between the unlabeled data and target task.

\textbf{(3) \method transfers useful knowledge from unlabeled data by data augmentation.} 
\method outperforms the best baseline relatively by +3.9\%, +13.4\%, and +10.2\% when there are only 1,210, 513, and 4,303 training graphs on \bace, \freesolv, and \glass, respectively. Compared to the self-supervised baselines, the improvement from \method is more significant, so the knowledge transfer is more effective. For example, on \freesolv and \oxygen, \method performs better than the best self-supervised baselines relatively by +45.8\% and +8.0\%, respectively. On regression tasks that involve knowledge transfer across domains (\eg from molecules to polymers), \method reduces MAE relatively by 1.9\%~$\sim$~10.2\% compared to the best baseline. All these results demonstrate the outstanding performance of task-specific data augmentation in \method.

\subsection{Ablation Studies and Performance Analysis}
\textbf{Comprehensive ablation studies:}
In~\cref{tab:ablation finetune}, we investigate how the task-related objectives in~\cref{eq:fine-tune objective} impact the performance of \method.
First, \method outperforms the top baseline even if the two task-related optimization objectives are disabled.
This is because \method generates new training examples based on original labeled graphs: the data augmentation has already improved the diversity of the training dataset a little bit.
Second, adding the objective $\I_1$ further improves the performance by encouraging the generation of diverse examples, because it minimizes the similarity between the original graph and augmented graph in the graph space.
Third, we receive the best performance of \method when it combines $\I_1$ and $\I_2$ objectives to generate task-related and diverse augmented graphs.
When we change the unlabeled data source from QM9 to the ZINC dataset from~\citep{jo2022score}, similar observations confirm the necessity of the task-related objectives.

\begin{table}[t!]
    \renewcommand{\arraystretch}{1.2}
    \renewcommand{\tabcolsep}{0.5mm}
    \vspace{-0.1in}
    \caption{\small Comprehensive ablation studies for \method on tasks \bace, \sider, \freesolv, and \oxygen. Objectives include minimizing $\I_1(G^\prime,G)$ and/or maximizing $\I_2(G^\prime, y)$.}
    \centering
    \begin{adjustbox}{width=0.55\textwidth}
    \begin{tabular}{@{}l|l|c|c|cc|cc@{}}
    \toprule
     \multicolumn{2}{c|}{} & \multicolumn{2}{c|}{Objectives} & \multicolumn{2}{c|}{Classification} & \multicolumn{2}{c}{Regression} \\
     \cmidrule{3-8}
      \multicolumn{2}{c|}{}& $\I_1(G^\prime,G)$ & $\I_2(G^\prime, y)$ & BACE & SIDER & FreeSolv & O$_2$Perm \\
    \midrule
    \multicolumn{4}{c|}{Top Baseline Method} & 82.4{\scriptsize (2.4)} & 60.8{\scriptsize (1.2)} & 1.547{\scriptsize (0.082)} &  177.7{\scriptsize (60.7)} \\
    \midrule
    \multirow{9}{*}{\rotatebox{90}{Unlabeled Data Sources}} &
    \multirow{4}{*}{\rotatebox{90}{ {QM9}}} 
    & \xmark  & \xmark
    & 84.4{\scriptsize (2.6)} & 63.7{\scriptsize (0.3)} & 1.473{\scriptsize (0.192)} & 177.4{\scriptsize (27.3)} \\
    && \cmark  & \xmark
    & 85.2{\scriptsize (1.3)} & 63.7{\scriptsize (0.2)} & 1.415{\scriptsize (0.145)} & 171.4{\scriptsize (14.0)} \\
    && \xmark  & \cmark
    & 84.7{\scriptsize (1.8)} & 63.8{\scriptsize (0.5)} & 1.344{\scriptsize (0.096)} & 172.6{\scriptsize (32.9)} \\
    && \cmark  & \cmark 
    & \textbf{85.6}{\scriptsize (0.6)} & \textbf{63.9}{\scriptsize (0.3)} & \textbf{1.339}{\scriptsize (0.075)} & \textbf{165.6}{\scriptsize (24.3)} \\
    \cmidrule{2-8}
    & \multirow{4}{*}{\rotatebox{90}{{ZINC}}} 
    & \xmark  & \xmark
    & 82.8{\scriptsize (1.8)} & 63.5{\scriptsize (0.7)} & 1.524{\scriptsize (0.219)} & 175.5{\scriptsize (11.9)} \\
    && \cmark  & \xmark
    & 83.3{\scriptsize (2.2)} & {63.5}{\scriptsize (0.7)} & 1.455{\scriptsize (0.207)} & 172.4{\scriptsize (60.8)} \\
    && \xmark  & \cmark
    & 84.3{\scriptsize (0.6)} & {63.5}{\scriptsize (0.6)} & 1.514{\scriptsize (0.214)} & 171.5{\scriptsize (26.0)} \\
    && \cmark  & \cmark 
    & 84.9{\scriptsize (0.4)} & {63.7}{\scriptsize (0.7)} & 1.408{\scriptsize (0.092)} & 169.3{\scriptsize (15.3)} \\
    \bottomrule
    \end{tabular}
    \end{adjustbox}
    \label{tab:ablation finetune}
\end{table}

\begin{figure*}[t]
    \centering
    \begin{subfigure}{\textwidth}
        \centering
        \includegraphics[width=0.98\linewidth]{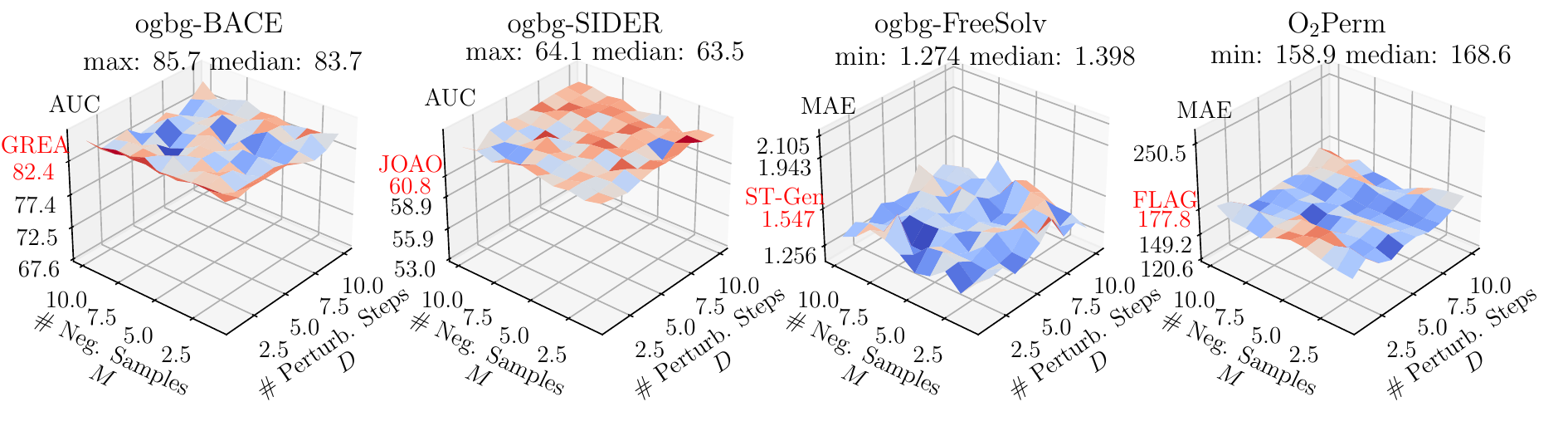} \label{fig:sensi d and m}
    \end{subfigure}%
    \hfill
    \begin{subfigure}{\textwidth}
        \centering
        \includegraphics[width=0.98\linewidth]{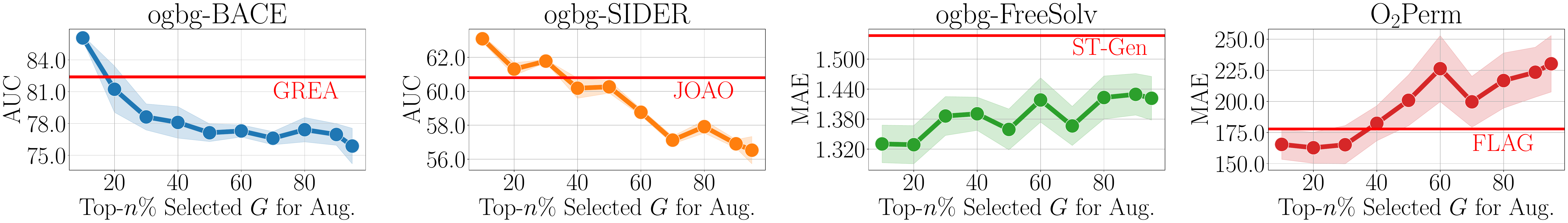}\label{fig:sensi topn}
    \end{subfigure}%
    \caption{Effect of hyper-parameters, including the number of perturbation steps $D \in [1,10]$, the number of negative graphs $M \in [1,10]$, and top-$n$~\% labeled graphs whose labels are predicted the most accurately and that are selected for data augmentation, where $n \in [10,100]$.}
    \label{fig:sensi all}
\end{figure*}

\textbf{Effect of hyper-parameters:} The impacts of three hyper-parameters of \method are studied: the number of perturbation steps $D$, the number of negative samples $M$ in~\cref{eq:upper bound infonce}, and the number of augmented graphs in each iteration (\ie top-$n$~\% selected graph for augmentation). Results from~\cref{fig:sensi all} show that \method is robust to a wide range of $D$ and $M$ valued from 0 to 10. They suggest that $D$ and $M$ can be set as 5 in most cases. As for the number of the augmented graphs in each iteration, results show that noisy graphs are often created when $n$ is higher than 30\%, because the predictor cannot effectively guide the data augmentation for those labeled graphs whose labels are hard to predict. So, 10\% is suggested as the default of top-$n$\%.

\begin{figure}[t]
\centering
\begin{minipage}[c]{0.61\textwidth}
    \includegraphics[width=0.95\linewidth]{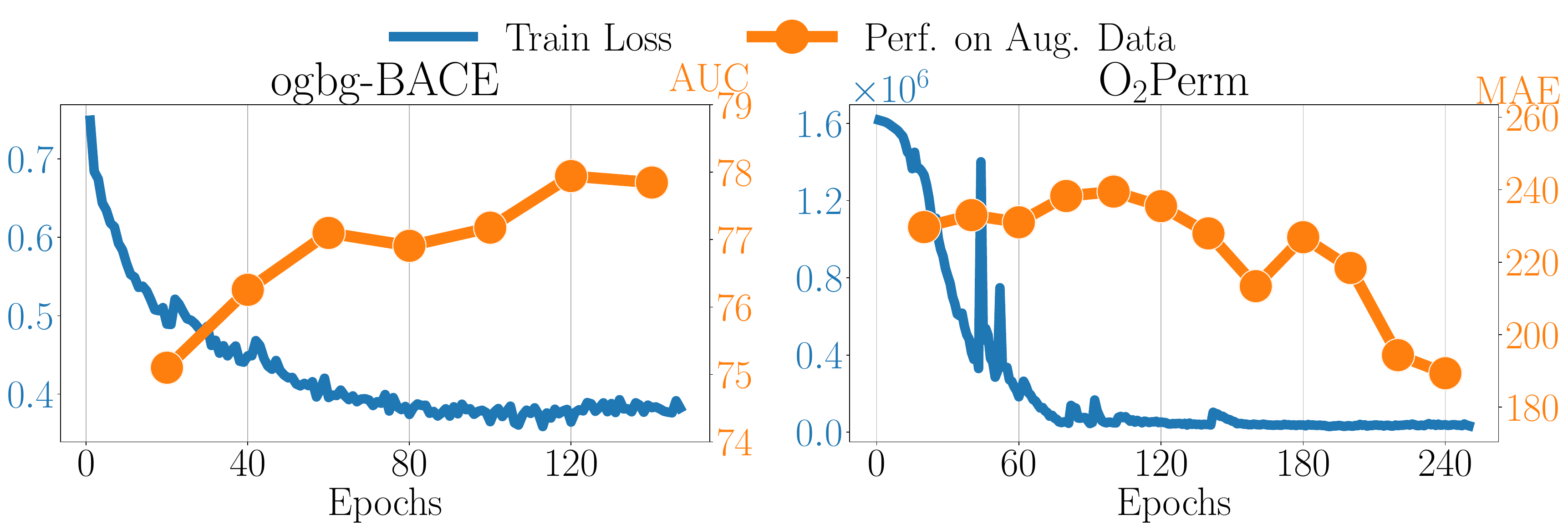}
\end{minipage}\hfill
\begin{minipage}[c]{0.38\textwidth}
    \caption{Data augmentation and model training mutually enhance each other over epochs. The predictor is saved every 20 epochs to guide the generation of augmented graphs. The performance of \gin trained on these augmented graphs reflects the quality of the augmented data.}
    \label{fig:mutual enhance}
\end{minipage}
\end{figure}

\textbf{Iterative process:} 
\cref{fig:mutual enhance} investigates the relationship between the quality of augmented graphs and the accuracy of property prediction models.
We save a predictor checkpoint every 20 epochs to guide the generation of the augmented examples.
We evaluate the quality of augmented graphs by using them to train \gin and report AUC/MAE. The data augmentation gradually decreases the training loss of property prediction. On the other hand, the increased \gin performance indicates that the quality of augmented examples is also improved over epochs. The data augmentation and predictor training mutual enhance each other.

\subsection{Interpretability of Visible Knowledge Transfer}
\begin{figure}[t]
    \centering
    \vspace{-0.25in}
    \begin{minipage}[c]{0.61\textwidth}
    \includegraphics[width=\linewidth]{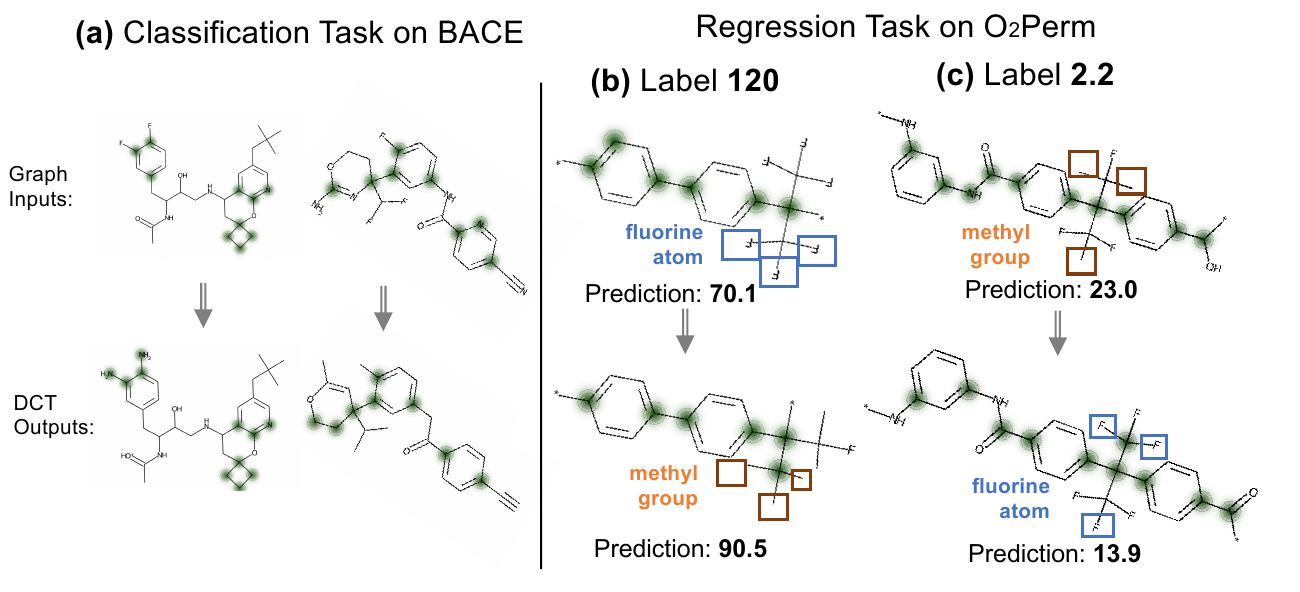}
    \end{minipage}\hfill
    \begin{minipage}[c]{0.38\textwidth}
    \caption{Case studies of augmented graphs. The green highlighted subgraphs are from \gin with top-k pooling. Examples show that the augmented graphs from~\method preserve the core structures of original graphs. Key concepts in the unlabeled graphs like chemical validity are transferred to downstream tasks. Domain knowledge such as the relationship between the permeability and the fluorine atom/methyl group is captured to guide task-specific generation. 
    }
    \label{fig:visual}
    \end{minipage}
    \vspace{-0.15in}
\end{figure}

\vspace{-0.09in}
Knowledge transfer by data augmentation gives visible examples, allowing us to study what is learned. We visualize a few augmented graphs in \method using \bace and \oxygen. We adapt top-k pooling~\citep{knyazev2019understanding} to select the subgraphs that \gin used for prediction. The selected subgraphs are highlighted in green in~\cref{fig:visual}. The three examples show that \emph{the augmented graphs can identify and preserve the core structures} that \gin uses to predict property values. These augmented graphs are chemically valid, showing that \emph{concepts such as some chemical rules from the unlabeled graphs are successfully transferred to downstream tasks}. More results are in~\cref{add:sec:chemical_valid}. Regarding task-specific knowledge, it is known that the fluorine atom and the methyl group are usually negatively and positively correlated to the permeability, respectively~\citep{park2003gas,corrado2020macromolecular}.
The augmented examples show that \emph{\method captures this domain knowledge with the task-related objectives}. In example (b), \method replaces most of the fluorine atoms with the methyl groups. It encourages \gin to learn the positive relationship between the methyl group and the permeability so that \gin predicts a high label value. In example (c), \method replaces the methyl groups with fluorine atoms. It encourages \gin to learn the negative relationship between the fluorine atom and the permeability so that \gin predicts a low label value.

\section{Related Work}
\label{sec:related}
\subsection{Graph Property Prediction}

Graph neural networks (GNNs)~\citep{kipf2017semi,xu2018powerful} are commonly used for graph property prediction in chemistry and polymer informatics tasks~\citep{otsuka2011polyinfo,hu2020open,zhou2022jointly}.
However, it is hard to annotate enough labels in these domains. 
Recent work used \textit{self-supervised tasks} such as node attribute prediction and graph structure prediction~\citep{hu2019strategies,you2021graph,kim2022graph} to pre-train architecture-fixed GNNs. \citet{sun2022does} observed that the existing methods might fail to transfer knowledge from unlabeled graph data. Flexible GNN architectures for downstream tasks would be desirable.

\textit{Graph data augmentation} (GDA) methods do not restrict GNN architecture choices to improve prediction accuracy~\citep{trivedianalyzing,zhao2022learning,zhao2022graph,ding2022data}.
They learn to create new examples that preserve the properties of original graphs~\citep{liu2022local,liu2022graph,kong2022robust,han2022g,luo2022automated}.
However, they purely manipulate labeled examples and thus \textit{cannot utilize the knowledge in unlabeled graphs}.
Our \method combines the knowledge from the unlabeled dataset and the labeled task dataset. It creates label-preserved graph examples with the knowledge transferred from the unlabeled data. It allows the GNN models to have flexible architectures.

\subsection{Learning from Unlabeled Data}

\textit{Pre-training on self-supervised tasks} such as masked image modeling and autoregressive text generation is effective for large language and vision models~\citep{brown2020language,he2022masked}. However, the hand-crafted self-supervised tasks could hardly help models learn useful knowledge from unlabeled graphs \emph{due to the gap between these label-agnostic tasks and the downstream prediction tasks} towards drug discovery and material discovery~\citep{sun2021mocl,kim2022graph,inae2023motif}.
A universal self-supervised task to learn from the unlabeled graphs remains under-explored~\citep{sun2022does,trivedianalyzing}.

\textit{Semi-supervised learning} assumes that unlabeled and labeled data are from the same source~\citep{liu2023semi}. The learning objective in the latent space is usually mutual information maximization that encourages similarity between the representations of unlabeled and labeled graphs~\citep{sun2019infograph}. However, \textit{the distributions of the unlabeled and labeled data could be very different} due to the different types of sources~\citep{hu2019strategies}, leading to negative impacts on the property prediction on the labeled graphs.
\textit{Self-training}, as a specific type of semi-supervised learning method, selects the unlabeled graphs of confidently predictable labels and assigns pseudo-labels for them~\citep{lee2013pseudo,iscen2019label}. Many studies have explored improving uncertainty estimation~\citep{gal2016dropout,tagasovska2019single,amini2020deep} to help the model filter out noise for reliable pseudo-labels. Recently, pseudo-labels have been applied in imbalanced learning~\citep{liu2023semi} and representation learning~\citep{ghiasi2021multi}. However, self-training is restricted to confidently predictable labels and may ignore the huge number of any other unlabeled graphs~\citep{huang2022uncertainty}. Therefore, it cannot fully utilize the knowledge in the unlabeled graphs. 

In contrast, our \method employs a diffusion model to extract knowledge (as the diffusion and reverse processes) from \textit{all the unlabeled graphs}. \method represents the knowledge as task-specific labeled examples to augment the target dataset, instead of uninterpretable pre-trained model parameters. We note that self- or semi-supervised learning does not conflict with \method, and we leave their combinations for future work.

\subsection{Diffusion Models on Graphs}
Recent works have improved the diffusion models on graphs~\citep{niu2020permutation,jo2022score,vignac2022digress,kong2023autoregressive,chen2023efficient}. EDP-GNN~\citep{niu2020permutation} employed score matching for permutation-invariant graph data distribution. GDSS~\citep{jo2022score} extended the continuous-time framework [6] to model node-edge joint distribution. DiGress~\citep{vignac2022digress} used the transition matrix to preserve the discrete natures of the graph structure. GraphARM~\citep{kong2023autoregressive} introduced a node-absorbing autoregressive diffusion process. EDGE~\citep{chen2023efficient} focused on efficiently generating larger graphs. Instead of improving the generation performance of the diffusion model, our model builds on score-based diffusion models~\citep{jo2022score,song2020score} for predictive tasks, \ie, graph classification and graph regression.

\section{Conclusion}
\label{sec:conclusion}
In this work, we made the first attempt to transfer minimal sufficient knowledge from unlabeled graphs by data augmentation. We proposed a data-centric framework to use the diffusion model trained on the unlabeled graphs and use two task-related objectives to generate task-specific augmented graphs. Experiments demonstrated the performance of the proposed framework through visible augmented examples. It is better than self-supervised learning, self-training, and graph data augmentation methods on as many as 15 tasks.

\begin{ack}
This work was supported by NSF IIS-2142827, IIS-2146761, IIS-2234058, CBET-2102592, and ONR N00014-22-1-2507.
\end{ack}

\bibliographystyle{plainnat}
\bibliography{ref}

\newpage
\appendix
\onecolumn
\section{Additional Related Work on Data-Centric Approach}\label{add:sec:related}
\paragraph{Data Augmentation}
Data augmentation creates new examples with preserved labels but uses no unlabeled data~\citep{shorten2019survey,kashefi2020quantifying,balestriero2022effects}. Examples of heuristic data augmentation techniques include flipping, distorting, and rotating images~\citep{shorten2019survey}, using lexical substitution, inserting words, and shuffling sentences in texts~\citep{kashefi2020quantifying}, and deleting nodes and dropping edges in graphs~\citep{zhao2021data,zhao2022graph}. While human knowledge can be used to improve data diversity and reduce over-fitting in heuristic methods, it is difficult to use a single heuristic method to preserve the different labels for different tasks~\citep{balestriero2022effects, cubuk2019autoaugment}. So, automated augmentation~\citep{cubuk2019autoaugment} learned from data to search for the best policy to combine a bunch of predefined heuristic augmentations. Generation models~\citep{antoniou2017data,bowles2018gan,han2022g} create in-class examples. Other learning ideas such as \textsc{FATTEN}~\citep{liu2018feature} and \textsc{GREA}~\citep{liu2022graph} learned to split the latent space for data augmentation. However, learning and augmenting from insufficient labels at the same time may limit the diversity of new examples and cause over-fitting. \method leverages unlabeled data to avoid them.

\paragraph{Relationship between Data-Centric Approaches} 
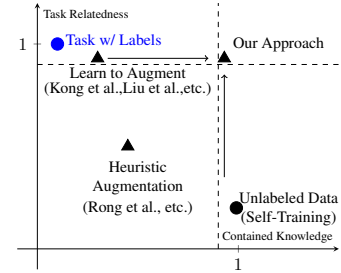
\begin{wrapfigure}{r}{5.5cm}
\caption{Qualitative relationship of graphs from different data-centric approach on the task relatedness and contained knowledge.}\label{fig: info relationship}
\resizebox{.8\linewidth}{!}{\begin{tikzpicture}
    \begin{axis}
    [
        legend pos= north east,
        axis lines = center,
        domain=-1:2,
        xlabel={\scriptsize Contained Knowledge},
        xmin=-0.1,
        xmax=1.5,
        xtick={0, 1},
        ylabel={\scriptsize Task Relatedness},
        ymin=-0.1,
        ytick={0, 1},
        ymax=1.2,
    ]
    \addplot[only marks, color=black, mark=triangle*, mark size=4pt] coordinates { (0.93,0.93) };
    \node[] at (axis cs: 1.2, 1.0) {\textcolor{black}{\small Our Approach}};

    \addplot[only marks, color=black, mark=triangle*,mark size=4pt] coordinates {(0.3,0.93)};
    \node[] at (axis cs: 0.45,0.85) {\textcolor{black}{\small Learn to Augment}};
    \node[] at (axis cs: 0.45,0.78) {\small (\citeauthor{kong2022robust},\citeauthor{liu2022graph},etc.)};
    
    \addplot[only marks, color=black, mark=triangle*,mark size=4pt]
        coordinates {(0.45,0.5) };
    \node[] at (axis cs: 0.5,0.4) {\textcolor{black}{\small Heuristic}};
    \node[] at (axis cs: 0.5,0.3) {\textcolor{black}{\small Augmentation}};
    \node[] at (axis cs: 0.5,0.2) {\textcolor{black}{\small (\citeauthor{rong2019dropedge}, etc.)}};
    
    \addplot[only marks, color=blue, mark=*, mark size=3.5pt]
        coordinates {(0.1,1)};
    \node[] at (axis cs: 0.38,1.02) {\textcolor{blue}{\small Task w/ Labels}};

    \addplot[only marks, color=black, mark=*, mark size=3.5pt]
        coordinates {(0.99,0.2) };
    \node[] at (axis cs: 1.25,0.25) {\textcolor{black}{\small Unlabeled Data}};
    \node[] at (axis cs: 1.25,0.15) {\textcolor{black}{\footnotesize (Self-Training)}};
    
    \draw [dashed] (0.9,0) -- (0.9,1.2);
    \draw [dashed] (0,0.9) -- (1.5,0.9);
    \draw[->](axis cs: 0.35,0.93)--(axis cs: 0.85,0.93);
    \draw[->](axis cs: 0.93,0.35)--(axis cs: 0.93,0.85);
    \end{axis}
\end{tikzpicture}}
\end{wrapfigure} 
As presented in~\cref{fig: info relationship}, perturb edges, delete nodes and mask attributes~\citep{rong2019dropedge,trivedianalyzing} for graphs are some heuristic ways for data augmentation. The augmented knowledge from them is mainly controlled by human prior knowledge on the perturbations and it often fails to be close to the task, \ie, random perturbations hardly preserve labels for the augmented graphs. The learning to augment approaches learn from labeled graphs to perturb graph structures~\citep{luo2022automated}, to estimate graphons for different classes~\citep{han2022g}, or to split the latent space for augmentation~\citep{liu2022graph}. Although these approaches could preserve labels for the augmented graphs, they introduce less extra knowledge to improve the model prediction. In summary, graph data augmentation is effective in expanding knowledge for limited labels, but it makes no use of unlabeled graphs. Besides, the diversity and richness of the domain knowledge from augmented graphs are far from that contained in a large number of unlabeled graphs. To learn from unlabeled graphs, data-centric approaches like the self-training is assumed to be useful when the unlabeled and labeled data are from the same source. It is less studied when we have a single unified unlabeled source for different tasks.

\section{Additional Method Details}
\subsection{Upper bounding the mutual information}
In~\cref{eq:upper bound infonce}, we use a leave-one-out variant of InfoNCE ($\I_\text{bound}$) to derive the upper bound of mutual information. We summarize the derivation~\citep{poole2019variational} here.
\begin{equation}\label{eq:add derivation mi upper bound}
    \begin{split}
        \mathcal{I}_1(G^\prime;G) & = \mathbb{E}_{p (G, G^\prime)} \left[ \operatorname{log} \frac{p(G^\prime|G)}{p(G^\prime)} \right] \\
        & = \mathbb{E}_{p (G, G^\prime)} \left[ \operatorname{log} \frac{p(G^\prime|G) q(G^\prime)}{q(G^\prime) p(G^\prime)} \right] \\
        & = \mathbb{E}_{p (G, G^\prime)} \left[ \operatorname{log} \frac{p(G^\prime|G)}{q(G^\prime)} \right] - \operatorname{KL} (p(G^\prime) || q(G^\prime) ) \\
        & \leq \mathbb{E}_{p (G, G^\prime)} \left[ \operatorname{log} \frac{p(G^\prime|G)}{q(G^\prime)} \right]
    \end{split}    
\end{equation}
The intractable upper bound is minimized when the variational approximation $q(G^\prime)$ matches the true marginal $p(G^\prime)$~\citep{poole2019variational}. 
For each $G_i$, its augmented output $G_i^\prime$, and $M-1$ negative examples with different labels, we could approximate $q(G_i^\prime) = \frac{1}{M-1} \sum_{j \neq i} p(G_i^\prime | G_j)$. So, we have
\begin{equation}
    \begin{split}
        \mathcal{I}_1(G^\prime_i, G_i) 
        & \leq  \operatorname{log} \frac{p(G_i^\prime| G_i)}{\frac{1}{K-1} \sum_{j=1, j\neq i}^M p(G_i^\prime | G_j)}
        \\ & = \operatorname{log} \frac{p(G_i^\prime| G_i)}{\sum_{j=1, j\neq i}^M p(G_i^\prime | G_j)} + \operatorname{log} (M-1)
        \\ & = \I_\text{bound} (G_i^\prime; G_i) + \text{constant}
    \end{split}
\end{equation}

\subsection{Extraction of Statistical Features on Graphs}\label{add:sec:raw feature}
For each molecule and polymer graph, we concatenate the following vectors or values for statistical feature extraction.
\begin{compactitem}
    \item the sum of the degree in the graph;
    \item the vector indicating the distribution of atom types;
    \item the vector containing the maximum, minimum and mean values of atoms weights in a molecule or polymer;
    \item the vector containing the maximum, minimum, and mean values of bond valence.
\end{compactitem}
For each protein-protein interaction ego-graph in the biology field, we use the sorted vector of node degree distribution in the graph as the statistical features.

\subsection{Technical Details for Graph Data Augmentation with Diffusion Model}\label{add:method:tech}

\paragraph{The Lookup Table from Atom Type to Node Embedding Space}
Given a graph $G$, we assume the node feature matrix on the graph is $\mathbf{X} \in \mathbb{R}^{n \times F_\text{n}}$, where $n$ is the number of nodes. The edge feature matrix is $\mathbf{E} \in \mathbb{R}^{m \times F_\text{e}}$, where $m$ is the number of edges. There are two ways for $G$ to represent the graph structure in practice. We can use either the dense adjacency matrix $\mathbf{A} \in \mathbb{R}^{n \times n}$ or sparse edge index $\mathbf{I}_e \in \mathbb{R}^{2 \times m}$. The diffusion model~\citep{jo2022score} on graphs prefers the former, which is more straightforward for graph generations. The prediction model prefers the latter because of its flexibility, and less computational cost and time. The transformation between two types of graph structure representation takes additional time. Particularly for molecular graphs, the node features used for generation (one-hot encoding of the atom type) and for prediction (see the official package of OGBG~\footnote{\url{https://github.com/snap-stanford/ogb/blob/master/ogb/utils/features.py}} for details) are different, which introduces extra time to process the graph data. For details, we (1) first need to extract discrete node attributes given the atom type and its neighborhoods; (2) we then need to use an embedding table to embed node attributes in a continuous embedding space; (3) the embedding features of nodes with their graph structure are inputted into the graph neural networks to get the latent representation for nodes. The reverse process for data augmentation in \method may need to repeatedly process graph data with steps (1) and (2). It introduces additional time. To address these technical problems, we build up a lookup table to directly map the atom type to the node embedding. We average the node attributes for the same type of node within the batch. We then use the continuous node attributes as weights to average the corresponding node embedding according to the table.

\paragraph{Instantiations of SDE on Graphs} According to~\citet{song2020score}, we use the Variance Exploding (VE) SDE for the diffusion process. Given the minimal noise $\sigma_{\min}$ and the maximal noise $\sigma_{\max}$, the VE SDE is:
\begin{equation}
\mathrm{d} G =\sigma_{\min }\left(\frac{\sigma_{\max }}{\sigma_{\min }}\right)^t \sqrt{2 \log \frac{\sigma_{\max }}{\sigma_{\min }}} \mathrm{d} \mathbf{w}, \quad t \in(0,1]
\end{equation}
The perturbation kernel is derived~\citep{song2020score} as:
\begin{equation}
p_{0 t}(G^{(t)} \mid G^{(0)})=\mathcal{N}\left(G^{(t)} ; G^{(0)}, \sigma_{\min }^2\left(\frac{\sigma_{\max }}{\sigma_{\min }}\right)^{2 t} \mathbf{I}\right), \quad t \in(0,1]
\end{equation}

On graphs, we follow~\citet{jo2022score} to separate the perturbation of adjacency matrix and node features:
\begin{equation}\label{eq:appendix:perturbation kernel}
    p_{0 t}(G^{(t)} \mid G^{(0)}) = p_{0 t}(\mathbf{A}^{(t)} \mid \mathbf{A}^{(0)}) p_{0 t}(\mathbf{X}^{(t)} \mid \mathbf{X}^{(0)}).
\end{equation}

\paragraph{The Sampling Algorithm in the Reverse Process for Graph Data Augmentation} We adapt the Predictor-Corrector (PC) samplers for the graph data augmentation in the reverse process. The algorithm is shown in~\cref{alg:diffusion augmentation}.
\begin{algorithm}[H]
   \caption{Diffusion-Based Graph Augmentation with PC Sampling}
   \small
   \label{alg:diffusion augmentation}
   \def\bfX{\mathbf{X}}
   \def\bfz{\mathbf{z}}
   \def\bfI{\mathbf{I}}
   \def\mclN{\mathcal{N}}
\begin{algorithmic}
    \STATE {\bfseries Input:} Graph $G$ with node feature $\mathbf{X}$ and adjacency matrix $\mathbf{A}$, the denoising function for node feature $\mathbf{s}_{\mathbf{X}}$ and adjacency matrix $\mathbf{s}_{\mathbf{A}}$, the fine-tune loss $\mathcal{L}_\textbf{aug}$, Lagevin MCMC step size $\beta$,  scaling coefficient $\epsilon_1$
    \STATE $\mathbf{A}^{(D)} \gets \mathbf{A} + \mathbf{z}_A $; \quad $\mathbf{z}_A \sim \mclN(\mathbf{0}, \bfI)$
    \STATE $\mathbf{X}^{(D)} \gets \mathbf{X} + \mathbf{z}_X $; \quad $\mathbf{z}_X \sim \mclN(\mathbf{0}, \bfI)$
    
   \FOR{$t=D-1$ {\bfseries to}  $0$}
    \STATE $\hat{G}_{(t+1)} \sim p_{0 t+1} ( \hat{G}_{(t+1)} |G^{(t+1)}) $ \COMMENT{inner-loop sampling with another PC sampler}
    
    \STATE $ \mathbf{S}_A = \frac{1}{2} \mathbf{s}_{\mathbf{A}}(G^{(t+1)}, t+1) - \frac{1}{2} 
    \alpha \nabla_{\mathbf{A}^{(t)}} \mathcal{L}_\textbf{aug}(\hat{G}_{(t+1)})$
    
    \STATE  $ \mathbf{S}_X = \frac{1}{2} \mathbf{s}_{\mathbf{X}}(G^{(t+1)}, t+1) - \frac{1}{2} 
    \alpha \nabla_{\mathbf{X}^{(t)}} \mathcal{L}_\textbf{aug}(\hat{G}_{(t+1)})$
    
    \STATE $\tilde{\mathbf{A}}^{(t)} \gets \mathbf{A}^{(t+1)} + g(t)^2 \mathbf{S}_A + g(t) \mathbf{z}_A $; \quad $\mathbf{z}_A \sim \mclN(\mathbf{0}, \bfI)$ \COMMENT{Predictor for adjacency matrix}
    \STATE $\tilde{\mathbf{X}}^{(t)} \gets \mathbf{X}^{(t+1)} +  g(t)^2 \mathbf{S}_X  + g(t) \mathbf{z}_X $; \quad $\mathbf{z}_X \sim \mclN(\mathbf{0}, \bfI)$ \COMMENT{Predictor for node features}
    \STATE $\mathbf{A}^{(t)} \gets \tilde{\mathbf{A}}^{(t)} + \frac{\beta}{2} \mathbf{S}_A + \epsilon_1 \sqrt{\beta} \mathbf{z}_A $; \quad $\mathbf{z}_A \sim \mclN(\mathbf{0}, \bfI)$ \COMMENT{Corrector for adjacency matrix}
    \STATE $\mathbf{X}^{(t)} \gets \tilde{\mathbf{X}}^{(t)} + \frac{\beta}{2} \mathbf{S}_X  + \epsilon_1 \sqrt{\beta} \mathbf{z}_X $; \quad $\mathbf{z}_X \sim \mclN(\mathbf{0}, \bfI)$ \COMMENT{Corrector for node features}
   \ENDFOR
   \STATE {return} $G^\prime  = ( \mathbf{A}^{(0)}, \mathbf{X}^{(0)})$
\end{algorithmic}
\end{algorithm}

\paragraph{The Algorithm of the Framework} The algorithm of the proposed data-centric knowledge transfer framework is presented in \cref{alg:dct stage 1} and \cref{alg:dct stage 2}. In detail, \cref{alg:dct stage 1} corresponds to \cref{sec:learning} and \cref{alg:dct stage 2} corresponds to \cref{sec:generating}.

\begin{minipage}{7cm}
\begin{algorithm}[H]
   \caption{The Data-Centric Knowledge Transfer Framework: Learning from Unlabeled Graphs}
   \label{alg:dct stage 1}
   \def\bfX{\mathbf{X}}
   \def\bfz{\mathbf{z}}
   \def\bfI{\mathbf{I}}
   \def\mclN{\mathcal{N}}
   \small
\begin{algorithmic}

    \STATE {\bfseries Input:} Given unlabeled graphs from the space $\mathcal{G}^{[U]}$, randomly initialized score models $\mathbf{s}_{\mathbf{X}}$ and $\mathbf{s}_{\mathbf{A}}$ for node feature and graph adjacency matrix, respectively, the total diffusion time step $T$.
    \WHILE{$\mathbf{s}_{\mathbf{X}}$ and $\mathbf{s}_{\mathbf{A}}$ not converged}
    \STATE Sample $G = (\mathbf{X}, \mathbf{A}) \in \mathcal{G}^{[U]}$
    \STATE Sample $t \in \operatorname{Uniform}({1, 2,\dots, T})$\
    \STATE Sample $\mathbf{z}_A \sim \mclN(\mathbf{0}, \bfI)$
    \STATE Sample $\mathbf{z}_X \sim \mclN(\mathbf{0}, \bfI)$
    \STATE Sample $\hat{G}$ with $t$, $\mathbf{z}_A$, $\mathbf{z}_X$ and~\cref{eq:appendix:perturbation kernel} 
    \STATE Optimize $\mathbf{s}_{\mathbf{A}}$ with the gradient: \\ \quad\quad   $\nabla \|\mathbf{z}_A - \mathbf{s}_{\mathbf{A}}(\hat{G}, t)  \|^2$
    \STATE Optimize $\mathbf{s}_{\mathbf{X}}$ with the gradient: \\ \quad\quad   $\nabla \|\mathbf{z}_X - \mathbf{s}_{\mathbf{X}}(\hat{G}, t)  \|^2$
    \ENDWHILE
\end{algorithmic}
\end{algorithm}
\end{minipage}
\hfil
\begin{minipage}{7.5cm}
\begin{algorithm}[H]
\vspace{0.02in}
   \caption{The Data-Centric Knowledge Transfer Framework: Generating Task-specific Labeled Graphs}
   \label{alg:dct stage 2}
   \def\bfX{\mathbf{X}}
   \def\bfz{\mathbf{z}}
   \def\bfI{\mathbf{I}}
   \def\mclN{\mathcal{N}}
   \small
\begin{algorithmic}
    \STATE {\bfseries Input:} Given task $k$ with the graph-label space ($\mathcal{G}, \mathcal{Y}$), a randomly initialized prediction model $f_\theta$, the well-trained score model  $\mathbf{s} = (\mathbf{s}_{\mathbf{X}}, \mathbf{s}_{\mathbf{A}})$, the training data set $\{G_i, y_i\}_i^{N_t}$, total training epoch $e$, the hyper-paramtere $n$
    \FOR{current epoch $e_i$ from 1 to $e$}
    \STATE Train $f_\theta$ on current training data $\{G_i, y_i\}_i^{N_t}$
    \IF{$e_i$ is divisible by the augmentation interval}
    \STATE Select $n$ graph-label pairs with the lowest training loss from $\{G_i, y_i\}_i^{N_t}$
    \STATE Get the augmented examples $\{G^\prime_i, y^\prime_i\}_i^{n}$ by~\cref{alg:diffusion augmentation} with the selected examples
    \STATE Update $\{G_i, y_i\}_i^{N_t}$ with $\{G^\prime_i, y^\prime_i\}_i^{n}$, \textit{e.g.,} add $\{G^\prime_i, y^\prime_i\}_i^{n}$ to $\{G_i, y_i\}_i^{N_t}$.
    \ENDIF
    \ENDFOR
\end{algorithmic}
\end{algorithm}
\end{minipage}

\section{Additional Experiments Set-ups}\label{sec: add exp setups}
We perform experiments on 15 datasets, including eight classification and seven regression tasks from chemistry, material science, and biology. We use Area under the ROC curve (AUC) for classification performance and mean absolute error (MAE) for regression.
\subsection{Molecule Classification and Regression Tasks}
Seven molecule classification and three molecule regression tasks are from open graph benchmark~\citep{hu2020open}. They were originally collected by MoleculeNet~\citep{wu2018moleculenet} and used to predict molecule properties. They include (1) inhibition to HIV virus replication in \hiv, (2) toxicological properties of 617 types in \toxcast, (3) toxicity measurements such as nuclear receptors and stress response in \toxt, (4) blood–brain barrier permeability in \bbbp, (5) inhibition to human $\beta$-secretase 1 in \bace, (6) FDA approval status or failed clinical trial in \clintox, (7) having drug side effects of 27 system organ classes in \sider, (8) predicting the property of lipophilicity in \lipo, (9) predicting the water solubility ($\log$ solubility in mols per litre) from chemical structures in \esol, (10) predicting the hydration free energy of molecules in water in \freesolv.
For all molecule datasets, we use the scaffold splitting procedure as the open graph benchmark adopted \citep{hu2020open}. It attempts to separate structurally different molecules into different subsets, which provides a more realistic estimate of model performance in experiments~\citep{wu2018moleculenet}.

\subsection{Polymer Regression Tasks}
Four polymer regression tasks include \glass, \melting, \thermal, and \oxygen. They are used to predict different polymer properties such as \emph{glass transition temperature} ($^\circ$C), \emph{melting temperature} ($^\circ$C), \emph{thermal conductivity} (W/mK) and \emph{oxygen permeability} (Barrer). \glass and \melting are collected from PolyInfo, which is the largest web-based polymer database \citep{otsuka2011polyinfo}. The \thermal dataset is from molecular dynamics simulation and is an extension from the dataset used in~\citep{ma2022machine}. The \oxygen dataset is created from the Membrane Society of Australasia portal,
consisting of a variety of gas permeability data \citep{thornton2012polymer}. Since a polymer is built from repeated units, researchers often use a single unit graph with polymerization points as polymer graphs to predict properties. Different from molecular graphs, two polymerization points are two special nodes (see ``$*$'' in \cref{fig:implementation}), indicating the polymerization of monomers \citep{cormack2004molecularly}. For all the polymer tasks, we randomly split by 60\%/10\%/30\% for training, validation, and test.

\subsection{Protein Classification Task}
An additional task is protein function prediction using protein-protein interaction graphs~\citep{hu2019strategies}. A node is a protein without attributes, an edge is a relation type between two proteins such as co-expression and co-occurrence. In our \method, we treat all the relations as the undirected edge without attributes.

\subsection{Baselines and Implementation}
When implementing \gin~\citep{xu2018powerful}, we tune its hyper-parameters for different tasks with an early stop on the validation set. We generally implement pre-training baselines following their own setting. For molecule and polymer property prediction and protein function prediction, the pre-trained \gin models with self-supervised tasks such as \edgepred, \attrmask, \contextpred in~\citep{hu2019strategies}, \infomax~\citep{velickovic2019deep} are available. So we directly use them. For other self-supervised methods, we implement their codes with default hyper-parameters. Following their settings, we use 2M ZINC15~\citep{sterling2015zinc} to pre-train \gin models for molecule and polymer property prediction. We use 306K unlabeled protein-protein interaction ego-networks~\citep{hu2019strategies} to pre-train the \gin for the downstream protein function property prediction. For self-training with real unlabeled graphs and \infograph~\citep{sun2019infograph}, we use 113K QM9~\citep{ramakrishnan2014quantum}. For self-training with generated unlabeled graphs, we train the diffusion model~\citep{jo2022score} on the real QM9 dataset and then produce the same number of generated unlabeled graphs. To train the diffusion model in our \method, we also use QM9~\citep{ramakrishnan2014quantum}.

\section{Additional Experiment Analysis}
\subsection{The Power of Diffusion Model to Learn from Unlabeled Graphs}
In~\cref{tab:ablation finetune}, when we replace the 133K QM9 with the 249K ZINC~\citep{jo2022score} to train the diffusion model, which nearly doubles the size of the unlabeled graphs and includes more atom types, we do not observe any additional improvement, and in some cases, even worse performance. It is possible because of the constraint of the current diffusion model's capacity to model the data distribution for a much larger number of more complex graphs. It encourages powerful generative models in the future, which could be directly used to benefit predictive models under the proposed framework.

\subsection{Chemical Validity of the Generated Graphs in Downstream Tasks}\label{add:sec:chemical_valid}
In \cref{fig:visual}, we show through some examples that concepts, such as certain chemical rules from the unlabeled graphs, are successfully transferred to downstream tasks. To further validate this point, we gathered 1,000 task-specific graphs generated in the intermediate steps on the tasks of \bace, \bbbp, \freesolv, and \oxygen. We then assessed the chemical validity of these graphs and observed that the validity is 92.8\%, 87.9\%, 97.4\%, and 62.1\%, respectively. Results show that transferring knowledge from pre-trained molecular data to target molecules yields relatively high chemical validity. However, the validity drops to 62\% when transferring knowledge from pre-trained molecular data to target polymer data. This finding indicates that the transferability of chemical rules becomes more challenging when the distribution gap between the pre-training data and downstream task data is larger.


\end{document}